\documentclass[letterpaper]{article} 
\usepackage{aaai2027}  
\usepackage[hyphens]{url}  
\usepackage{graphicx} 
\usepackage{bm}
\usepackage{natbib}  
\usepackage{caption} 
\usepackage{amsmath}
\usepackage{amssymb}
\usepackage{algorithm}
\usepackage{algorithmic}
\usepackage{multirow}
\usepackage[table]{xcolor}

\usepackage{newfloat}
\usepackage{listings}
\DeclareCaptionStyle{ruled}{labelfont=normalfont,labelsep=colon,strut=off} 
\floatstyle{ruled}
\newfloat{listing}{tb}{lst}{}
\floatname{listing}{Listing}

\usepackage{booktabs}

\title{GeoMFD: Continual Drone-View Geo-Localization with Geometry-Aware Adapter and Margin-Field Distillation}
\author {
    Zhongwei Chen\textsuperscript{\rm 1},
    Hai-jun Rong\textsuperscript{\rm 1},
    Tao Zhang\textsuperscript{\rm 1},
    Xianfeng Nie\textsuperscript{\rm 1},
    Xiangbao Zhang\textsuperscript{\rm 1}\\
    Guoqi Li\textsuperscript{\rm 2}\corresponding,
    Zhao-Xu Yang\textsuperscript{\rm 1}\corresponding
}
\affiliations {
    \textsuperscript{\rm 1}School of Aerospace Engineering, Xi'an Jiaotong University \\
    \textsuperscript{\rm 2}Institute of Automation, Chinese Academy of Sciences \\
ISChenawei@sut.xjtu.edu.cn,yangzhx@xjtu.edu.cn, guoqi.li@ia.ac.cn\\
    Code: \textcolor{magenta}{\url{https://github.com/ISChenawei/GeoMFD}}
}
\begin{document}
\makeatletter
\def\copyright@text{}
\makeatother

\maketitle

\begin{abstract}
Existing drone-view geo-localization (DVGL) methods are mainly developed under a static training paradigm, where models are optimized for fixed environments with all training data available in advance. However, this paradigm is difficult to extend to real-world deployment, where drones may encounter diverse environments and require multiple environment-specific models, resulting in additional storage and model-selection costs. Directly adapting a single model to new environments also risks distorting previously learned cross-view embedding geometry and causing forgetting. To address these challenges, we formalize the continual drone-view geo-localization (C-DVGL) setting and propose GeoMFD, a geometry-aware continual adaptation method for DVGL. GeoMFD combines a cold-start bootstrapping strategy (CBS), a geometry-aware adapter (Geo-Adapter), and margin-field distillation (MFD) to balance adaptation and cross-view geometry preservation. CBS initializes a stable embedding space, Geo-Adapter enables environment adaptation through controlled residual corrections, and MFD preserves similarity margins between positive pairs and hard negatives to alleviate cross-view geometry forgetting. Extensive experiments demonstrate that GeoMFD effectively mitigates forgetting and achieves competitive performance with environment-specific DVGL methods using a single continuously updated model.

\end{abstract}

\section{Introduction}


Drone-view geo-localization (DVGL) is a promising visual localization task for GNSS-denied environments, where a drone-view query is matched with its corresponding geo-referenced satellite image~\cite{shen2023mccg}. Recent methods have achieved substantial progress through advanced cross-view representation learning~\cite{lv2024direction,wang2025coarse}. However, most are developed under a closed-world assumption, where the model is trained and evaluated within a fixed target environment~\cite{wang2024multiple,song2026geobridge}. This setting limits practical deployment, as drones may operate across diverse environments, including urban areas and campuses under diverse daytime and nighttime conditions. Variations in illumination, viewpoint, flight altitude, and scene structure can cause substantial performance degradation when the model encounters unseen environments. Training a separate model for each environment is a straightforward alternative, but it increases storage costs, complicates onboard deployment, and requires an additional mechanism to select the appropriate model during operation.

Therefore, a more practical solution is to maintain a single DVGL model that can be progressively adapted to new environments, avoiding the storage and selection of multiple environment-specific models. Continual learning provides a natural framework for this goal, yet its direct application to DVGL remains challenging. Existing methods mainly address classification or same-modality retrieval, where forgetting is reflected in degraded category or identity discrimination \cite{zhao2026representation,wen2026prompt}. In contrast, DVGL depends on the relative geometry between drone-view and satellite-view embeddings. Sequential adaptation may bias the embedding space toward the current environment, reducing the margins between positive pairs and hard negatives and disrupting previously learned cross-view relationships. Motivated by this observation, we introduce continual drone-view geo-localization (C-DVGL), which extends standard retrieval between drone and satellite imagery to evolving environments. In this setting, a single model is progressively updated across environments while retaining the cross-view matching capability learned from previous ones. The central challenge is therefore to adapt the model to new environments without distorting the previously established cross-view embedding geometry.

\begin{figure*}[t]
  \centering
  \includegraphics[width=6.8in]{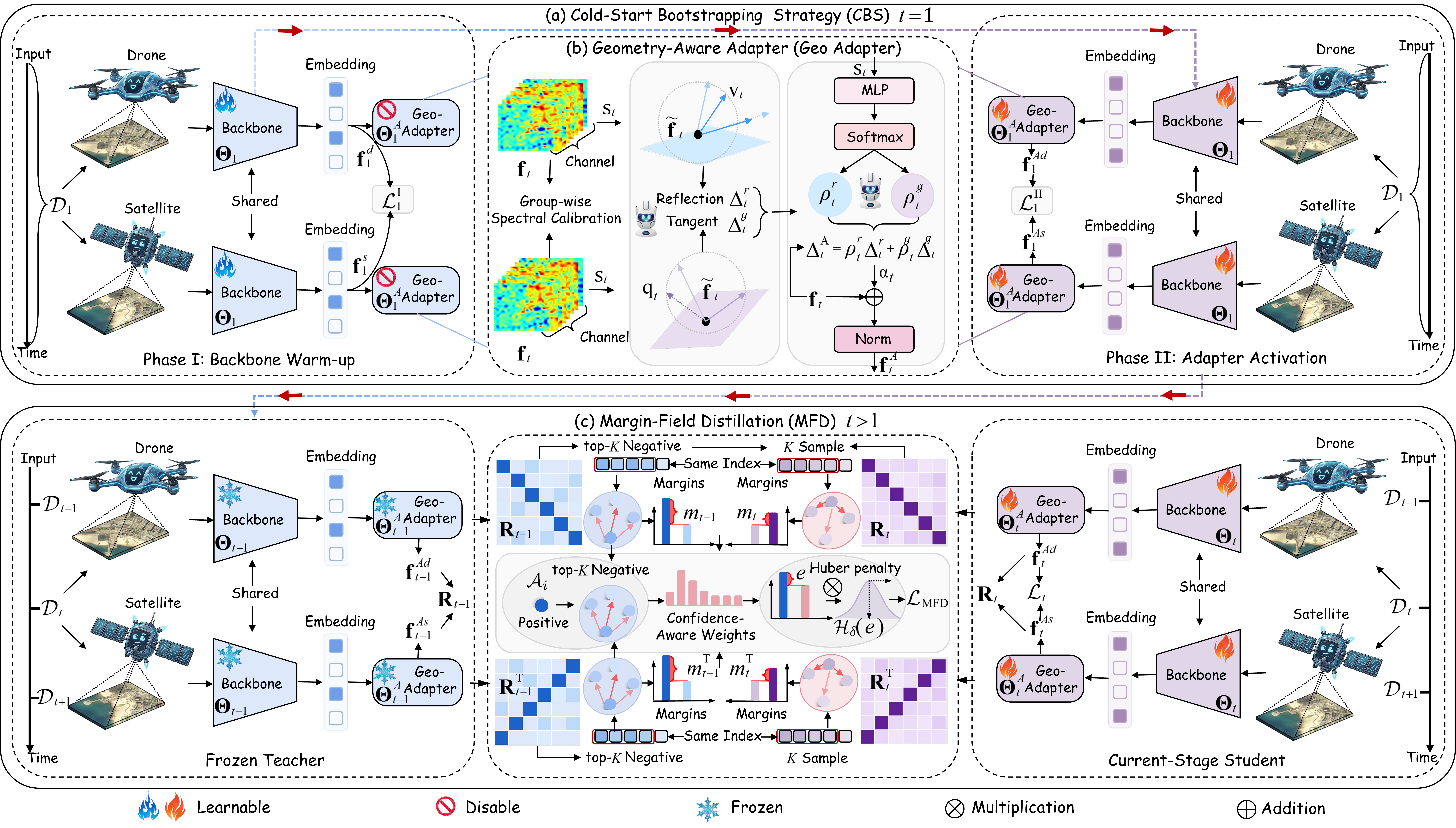}
    \caption{Pipeline Overview.
    Our method contains three key components. 
    (a) At the initial stage, cold-start bootstrapping warms up the shared backbone with the adapter disabled and then activates the adapter for stable residual learning.
    (b) The geometry-aware adapter performs view-gap-aware angular correction by combining spectral calibration, reflection residuals, tangent residuals, and adaptive gating. 
    (c) For later stages, margin-field distillation uses the frozen previous-stage model as a teacher to preserve positive-versus-hard-negative margins, mitigating forgetting while retaining plasticity for new environments.}
  \label{fig1}
\end{figure*}

To address this challenge, we propose GeoMFD, a geometry-aware continual adaptation method that balances environmental adaptation and cross-view geometry preservation. Specifically, as illustrated in Fig.~\ref{fig1}, GeoMFD consists of three complementary components, including a cold-start bootstrapping strategy (CBS), a geometry-aware adapter (Geo-Adapter), and margin-field distillation (MFD). CBS first constructs a stable initial cross-view embedding space, providing a reliable geometric foundation for subsequent continual adaptation. Based on this initialization, Geo-Adapter applies controlled residual corrections in the normalized embedding space, enabling adaptation to new environments while limiting undesirable distortion of the learned cross-view embedding geometry. Meanwhile, MFD preserves the relative similarity margins between positive pairs and hard negatives established in previous stages, thereby reducing geometry forgetting without retaining data from earlier environments. Extensive experiments demonstrate that GeoMFD effectively mitigates forgetting and achieves competitive performance with environment-specific models using a single continuously updated model. Our contributions are summarized as follows:
\begin{itemize}
\item We formalize the continual drone-view geo-localization (C-DVGL) setting, extending DVGL from learning in static environments to sequential environment adaptation.

\item We propose GeoMFD, a geometry-aware continual adaptation method consisting of CBS, Geo-Adapter, and MFD. CBS initializes a stable initial cross-view embedding space as the geometric foundation for subsequent adaptation. Geo-Adapter enables environment adaptation through controlled residual corrections and preserves the learned structure, and MFD maintains the relative similarity margins between positive pairs and hard negatives to prevent cross-view geometry degradation and catastrophic forgetting.

\item Extensive experiments under various continual task orders demonstrate that GeoMFD effectively mitigates forgetting and achieves competitive performance with environment-specific DVGL models while requiring only a single continuously updated model.
\end{itemize}

\section{Related Work}
\label{sec:related_work}

\subsection{Drone-View Geo-Localization}
\label{subsec:related_dvgl}

Drone-view geo-localization (DVGL) aims to retrieve the geo-referenced satellite image corresponding to a given drone-view query \cite{11540350}. Existing methods primarily focus on cross-view representation learning and
fine-grained feature alignment. LPN~\cite{wang2021each} extracts part-level features through local pattern partitioning, and FSRA~\cite{dai2021transformer} further selects salient regions to enhance region-level discrimination. TransFG~\cite{zhao2024transfg} captures long-range cross-view dependencies through Transformer-based feature modeling. CAMP~\cite{wu2024camp} combines contrastive attribute mining with position-aware partitioning, DAC~\cite{xia2024enhancing} improves cross-view consistency through domain alignment and scene constraints, and MEAN~\cite{chen2024multi} strengthens global-to-local associations via multi-level embedding alignment. Recent studies further extend DVGL by incorporating geometric structures and additional modalities. Geometry-aware methods exploit orientation priors, multi-view observations, or 3D structural cues to reduce the geometric discrepancy between oblique drone-views and overhead satellite images~\cite{li2025unsupervised,ji2025mmgeo,liu2026satellitefree}. 

Despite these advances, existing DVGL methods are mainly developed under a static training paradigm, where models are optimized for fixed target environments. Although effective for benchmark evaluation, this assumption limits real-world deployment, as adapting to new environments often requires retraining or maintaining multiple environment-specific models, increasing storage and deployment costs. In contrast, this work introduces continual learning into DVGL and proposes GeoMFD to adapt a single model to sequentially changing environments while preserving cross-view embedding geometry and performance.
\subsection{Continual Learning}
\label{subsec:related_cl}

Continual learning aims to update models with sequentially arriving data while mitigating catastrophic forgetting. In classification tasks, existing methods mainly preserve previous task knowledge through parameter regularization, exemplar replay, and knowledge distillation~\cite{kirkpatrick2017overcoming,buzzega2020dark,li2017learning}. Beyond classification, continual learning has been extended to retrieval scenarios, particularly lifelong person re-identification, where models need to adapt to new identities and domains while preserving previous retrieval ability. CLUDA-ReID~\cite{huang2022lifelong} balances domain adaptation and forgetting resistance, and KRC~\cite{yu2023lifelong} consolidates historical knowledge for continual retrieval. Recent works further explore embedding-level knowledge preservation in lifelong retrieval. LSTKC and LSTKC++~\cite{xu2024lstkc,11010188} model long- and short-term knowledge evolution, DKP~\cite{xu2024distribution} preserves distribution-aware prototypes, and DASK~\cite{xu2025dask} rehearses historical feature distributions through adaptive style kernels.
 
However, existing continual retrieval methods are mainly designed for identity-centric matching, where the objective is to preserve identity discrimination or retrieval consistency. In contrast, C-DVGL requires preserving the cross-view embedding geometry between heterogeneous drone and satellite views, which involves maintaining relative similarity structures rather than identity features. Therefore, directly applying existing retrieval continual learning strategies may not effectively prevent cross-view geometry degradation, motivating geometry-aware continual adaptation for DVGL.

\section{Methodology}

\textbf{Problem Definition}. 
In C-DVGL, training data are observed from a sequence of environments, and only the current stage data are available. Let $\mathcal{D}=\{\mathcal{D}_t\}_{t=1}^{T}$ denote the training sequence, where each $\mathcal{D}_t=(\mathcal{U}_t,\mathcal{V}_t)$ contains drone-view and satellite-view images annotated with location identities. A cross-view pair is positive if both images correspond to the same location. The final model is evaluated on unseen locations from all observed environments, aiming to adapt to new environments while preserving previous performance.

\noindent\textbf{Notation}. 
At stage $t$, the drone-view and satellite-view branches share the same backbone $\bm{\Theta}_t$ and geometry-aware adapter $\bm{\Theta}_t^A$. Given a drone-view image $u_i$ and a satellite-view image $v_j$, their backbone features are $\mathbf{f}_{t,i}^{d}=\bm{\Theta}_t(u_i)$ and $\mathbf{f}_{t,j}^{s}=\bm{\Theta}_t(v_j)$, and their adapted features are $\mathbf{f}_{t,i}^{Ad}=\bm{\Theta}_t^A(\mathbf{f}_{t,i}^{d})$ and $\mathbf{f}_{t,j}^{As}=\bm{\Theta}_t^A(\mathbf{f}_{t,j}^{s})$. $\bar{\mathbf{f}}$ denotes an $\ell_2$-normalized feature. For a mini-batch of $B$ matched pairs, we define the cross-view affinity matrix $\mathbf{R}_t\in\mathbb{R}^{B\times B}$ by $(\mathbf{R}_t)_{ij}=\beta_t(\bar{\mathbf{f}}_{t,i}^{Ad})^\top\bar{\mathbf{f}}_{t,j}^{As}$, where $\beta_t$ is the logit scale.

\subsection{Backbone and Geometry-Aware Adapter}
\label{subsec:adapter}

To enable environment adaptation while preserving the cross-view embedding geometry, GeoMFD combines a shared feature backbone with a lightweight geometry-aware adapter. Specifically, the drone and satellite branches share a ViT-B/16 backbone initialized with DINOv3 weights~\cite{simeoni2025dinov3}, and the adapter is attached to its output through a gated residual pathway to compensate for feature shifts. Since DVGL relies on cosine similarity between $\ell_2$-normalized features, retrieval is governed by angular geometry on the unit hypersphere. Given a backbone feature $\mathbf{f}_{t}\in\mathbb{R}^{C}$, the adapter first applies layer normalization and computes the mean absolute response of each channel group to form the spectral statistics $\mathbf{s}_t$. Then, a lightweight bottleneck maps $\mathbf{s}_t$ to group-wise scaling factors, which rescale the corresponding channels to obtain the calibrated feature $\widetilde{\mathbf{f}}_t$. The calibrated feature is then fed into two MLPs, $h_t^r(\cdot)$ and $h_t^g(\cdot)$, to predict a normalized reflection direction $\mathbf{v}_t$ and a local correction vector $\mathbf{q}_t$, respectively:
\begin{equation}
\mathbf{v}_t
=
h_t^r(\widetilde{\mathbf{f}}_t)
\,/\,
\left\|
h_t^r(\widetilde{\mathbf{f}}_t)
\right\|_2,
\quad
\mathbf{q}_{t}
=
h_t^g(\widetilde{\mathbf{f}}_t),
\label{eq:adapter_branches}
\end{equation}
where $\mathbf{v}_t$ defines a Householder-style reflection direction, while $\mathbf{q}_t$ provides a local correction direction. Based on these two directions, we construct a reflection residual and a tangent residual:
\begin{equation}
\Delta_{t}^{r}
=
\left(
\bar{\mathbf{f}}_{t}
-
2
\langle
\bar{\mathbf{f}}_{t},
\mathbf{v}_{t}
\rangle
\mathbf{v}_{t}
\right)
-
\bar{\mathbf{f}}_{t},
\quad
\Delta_{t}^{g}
=
\mathbf{q}_{t}
-
\langle
\mathbf{q}_{t},
\bar{\mathbf{f}}_{t}
\rangle
\bar{\mathbf{f}}_{t}.
\label{eq:geo_residuals}
\end{equation}
Here, $\bar{\mathbf{f}}_{t}=\mathbf{f}_{t}/\|\mathbf{f}_{t}\|_{2}$. $\Delta_t^r$ captures an angular correction through reflection, while $\Delta_t^g$ removes the radial component of $\mathbf{q}_t$ and performs correction in the tangent space of the unit hypersphere. The two residuals are adaptively fused by the gate:
\begin{equation}
\Delta_t^A(\mathbf{f}_{t})
=
\rho_t^r\Delta_t^r
+
\rho_t^g\Delta_t^g,
\label{eq:adapter_residual}
\end{equation}
where $[\rho_t^r,\rho_t^g]=\mathrm{softmax}(\psi_t(\mathbf{s}_{t}))$.
$\psi_t(\cdot)$ is a lightweight MLP that predicts two logits corresponding to the reflection residual $\Delta_t^r$ and the tangent correction residual $\Delta_t^g$. This allows each sample to adaptively balance global reflection and local tangent correction. Finally, the adapted feature is obtained as
\begin{equation}
\mathbf{f}_{t}^{A}
=
\bm{\Theta}_{t}^{A}(\mathbf{f}_{t})
=
\mathrm{Norm}
\left(
\bar{\mathbf{f}_{t}}
+
\alpha_t
\Delta_{t}^{A}(\mathbf{f}_{t})
\right),
\label{eq:adapter_output}
\end{equation}
where $\alpha_t$ is a learnable scalar initialized to a small value. The final normalization keeps the adapted feature on the unit hypersphere, and the small residual gate makes the adapter start from a near-identity mapping before gradually learning environment-specific angular corrections.

\subsection{Cold-Start Bootstrapping Strategy}
\label{subsec:cold_start}

At the first stage, directly activating the adapter may be unstable because the cross-view retrieval space is not yet calibrated. We thus adopt a two-phase cold-start bootstrapping strategy on $\mathcal{D}_1$, applied only at $t=1$. The backbone is first warmed up to build a basic cross-view embedding space, and the adapter is then activated for geometry-aware residual learning. To formulate the objectives of the two phases, let $\mathcal{B}_1=\{(\tilde{u}_b,\tilde{v}_b)\}_{b=1}^{B}$ denote a mini-batch of matched cross-view pairs. For a similarity matrix $\mathbf{S}\in\mathbb{R}^{B\times B}$, with diagonal positives and off-diagonal in-batch negatives, the row-wise contrastive loss is defined as
\begin{equation}
\mathcal{C}(\mathbf{S})
=
-\frac{1}{B}
\sum_{b=1}^{B}
\log
\frac{\exp(S_{bb})}
{\sum_{k=1}^{B}\exp(S_{bk})}.
\label{eq:row_nce}
\end{equation}
The symmetric cross-view contrastive objective is written as $\mathcal{J}(\mathbf{S})=\frac{1}{2}[\mathcal{C}(\mathbf{S})+\mathcal{C}(\mathbf{S}^{\top})]$, which performs contrastive learning in both directions.

\textbf{Phase I: Backbone warm-up.}
In the first phase, the adapter is disabled, and only the shared backbone $\bm{\Theta}_1$ is optimized. For each matched pair $(\tilde{u}_b,\tilde{v}_b)$, the backbone features are denoted as $\mathbf{f}_{1,b}^{d}=\bm{\Theta}_1(\tilde{u}_b)$ and $\mathbf{f}_{1,b}^{s}=\bm{\Theta}_1(\tilde{v}_b)$. We construct the backbone-level similarity matrix by $S_{bk}^{I}=\mathrm{sim}(\mathbf{f}_{1,b}^{d},\mathbf{f}_{1,k}^{s})/\tau_c$, where $\tau_c$ is the contrastive temperature. The Phase-I objective is $\mathcal{L}_{1}^{\mathrm{I}}=\mathcal{J}(\mathbf{S}^{I})$. After this phase, the warmed-up backbone is denoted as $\bar{\bm{\Theta}}_1$.

\begin{table*}[ht]
\centering
\small
\setlength{\tabcolsep}{3.5pt}
\begin{tabular}{l|l|cc|cc|cc|cc|cc|cc|c}
\hline
\multirow{2}{*}{Protocol} & \multirow{2}{*}{Method} 
& \multicolumn{2}{c|}{IR-VL328} 
& \multicolumn{2}{c|}{DenseUAV} 
& \multicolumn{2}{c|}{SUES-200} 
& \multicolumn{2}{c|}{C-RGBT} 
& \multicolumn{2}{c|}{U-1652} 
& \multicolumn{2}{c|}{Avg} 
& \multirow{2}{*}{Storage}\\
& & R@1 & AP & R@1 & AP & R@1 & AP & R@1 & AP & R@1 & AP & R@1 & AP &\\
\hline
\multicolumn{15}{c}{\textbf{Drone $\rightarrow$ Satellite}} \\
\hline

\multirow{6}{*}{Individual} 
&Sample4Geo & 68.19 & 73.43 & 85.54 & 67.24 & 92.60 & 94.00 & 89.33 & 91.46 &92.65 & 93.81 & 85.66 & 83.99 &1752 \\
&MEAN & 63.75 & 70.11 & \underline{86.61} & \underline{67.89} & 95.50 & 96.46 & 81.33 & 85.31 & 93.55 & 94.53 & 84.15 & 82.86 & 730.5 \\
&MFRGN& 67.81 & 73.25 & \textbf{93.47} & \textbf{80.12} & 95.85 &96.52 &88.00 & 90.50  & 94.33 & 95.24 & 87.89 & \underline{87.13} & 1869.5 \\
&CAMP & 67.04 & 72.60 & 84.47 & 66.02 & 95.40 & 96.38 & 85.33 & 88.03 & 94.46 & 95.38 & 85.34 & 83.68 &1828.5\\
&DAC & \underline{71.15} & \underline{75.56} & 85.67 & 68.03 &\underline{96.80} &\underline{97.54} & \underline{91.32} & \textbf{93.12} & \underline{94.67} & \underline{95.50} &\underline{87.92} & 85.95 &  1931 \\
&SURFNet & 65.19 & 70.87 &77.52 & 58.73 & 93.63 & 94.84  & 88.00 & 90.56  &94.57 & 95.49  &83.78 &82.10 &2124  \\\hline

\multirow{3}{*}{Continual} &DKP &22.42  &29.10  & 20.93 &11.41 & 48.64 &56.19 &40.74 &48.10 & 58.87 & 63.72 & 38.32 & 41.70 &409.0\\
&LSTKC++ &{31.25}  &{38.22}  & 17.16 &10.99 & {21.10} & {27.86} & 90.00 & 92.11 & {12.33} &{15.93} &{34.37} & {37.02} &{126.1}\\
&DASK &{22.45}  &{28.51}  & 14.84 &9.11 & {51.17} & {57.52} & 78.66 & 82.23 & {44.32} &{49.34} &{42.29} & {45.34} &\textbf{126.1}\\\hline
\multirow{1}{*}{Continual} &GeoMFD & \textbf{86.76} & \textbf{89.25} &73.70 & 58.85 & \textbf{99.17} & \textbf{99.36} & \textbf{91.33} & \underline{93.00}&\textbf{96.23} & \textbf{96.93} & \textbf{89.44} & \textbf{87.48}  & \underline{345.9}\\
\hline

\multicolumn{15}{c}{\textbf{Satellite $\rightarrow$ Drone}} \\
\hline
\multirow{6}{*}{Individual}
&Sample4Geo& 75.78 & 63.92 & \underline{20.45} & 18.54  & 97.50 & 93.63  & 88.00 & 90.65 & 95.14 & 91.39 & 75.37 & 71.63 &1752 \\
&MEAN& 68.75 & 57.36 & 20.33 & 18.26 & 97.50 & \underline{94.75} & 86.66  & 89.45  & 96.01 & 92.08 & 73.85 & 70.38 & 730.5 \\
&MFRGN & \underline{78.12} & 59.33 &\textbf{23.35} & \textbf{21.40} & \underline{98.75} & 94.42 & 84.00 & 87.48 & 96.15 & \underline{93.94} & {76.07} & 71.31& 1869.5 \\
&CAMP& 71.87 & 63.37 &19.56 & 17.71 & 96.25 & 93.36 & 83.33 & 86.42 & 96.15 & 92.72 & 73.43 & 70.72 &1828.5\\
&DAC& 75.00 & \underline{65.38} & 20.25 & \underline{18.55} &97.50 &94.06 & \textbf{93.33} &\textbf{94.71} & \underline{96.43} & 93.79 & \underline{76.50} & \underline{73.30} & 1931 \\
&SURFNet&71.88  &59.01  & 17.61 & 16.07 &95.00 & 90.75 &89.33 & 91.13 &95.72 &93.20 & 73.91 &70.03 &2124  \\\hline
\multirow{3}{*}{Continual} &DKP & 19.53 & 16.46 & 12.05 & 10.83 & 61.25 & 49.92 & 44.00 & 51.53 & 79.03 & 59.21 & 43.17 & 37.59 & 409.0\\
&LSTKC++ &{71.09}  &{22.41}  & 13.94 &12.26 & {36.25} & {17.53} & 91.33 & 92.94 & {42.36} &{10.85} &{50.99} & {31.20} &{126.1}\\
&DASK &{25.00}  &{10.35}  & 6.58 &6.33 & {52.50} & {42.39} &84.00 & 87.30 & {77.31} &{45.07} &49.08 &38.29 &\textbf{126.1}\\\hline 
\multirow{1}{*}{Continual} &GeoMFD &\textbf{90.62}  &\textbf{81.35}  & 17.68 &16.63 & \textbf{100.00} & \textbf{98.28} & \underline{92.00} & \underline{93.84} & \textbf{96.71} & \textbf{95.59} &\textbf{79.40} & \textbf{77.14} & \underline{345.9}\\\hline
\end{tabular}
\caption{Comparison of GeoMFD with state-of-the-art DVGL methods independently trained on each dataset and state-of-the-art continual retrieval methods adapted to the C-DVGL setting.
For SUES-200, results are reported at 150\,m.} 
\label{tab1}
\end{table*}

\begin{figure*}[ht]
  \centering
  \includegraphics[width=7in]{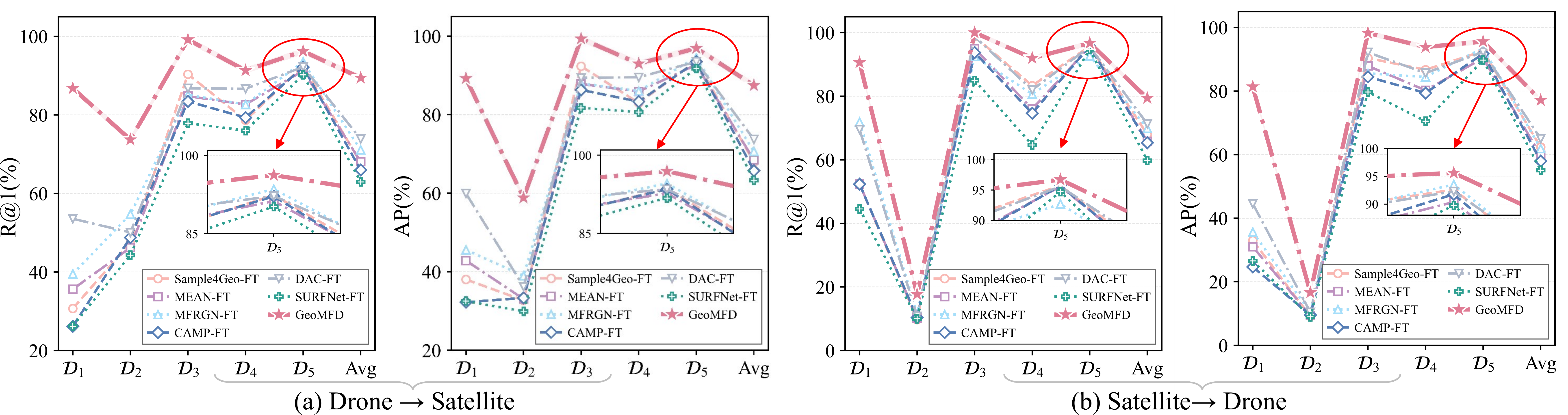}
    \caption{This figure complements Table~\ref{tab1} by comparing GeoMFD with state-of-the-art DVGL methods under sequential fine-tuning, denoted by the suffix ``-FT''. For example, DAC-FT refers to the sequentially fine-tuned version of DAC.}
  \label{fig2}
\end{figure*}

\textbf{Phase II: Adapter activation.}
In the second phase, the backbone $\bm{\Theta}_1$ is initialized from $\bar{\bm{\Theta}}_1$, and the geometry-aware adapter $\bm{\Theta}_1^{A}$ is activated. Both modules are jointly optimized. For each matched pair $(\tilde{u}_b,\tilde{v}_b)$, the adapted features are written as $\mathbf{f}_{1,b}^{Ad}=\bm{\Theta}_1^{A}(\bm{\Theta}_1(\tilde{u}_b))$ and $\mathbf{f}_{1,b}^{As}=\bm{\Theta}_1^{A}(\bm{\Theta}_1(\tilde{v}_b))$. We construct the adapter-level similarity matrix by $S_{bk}^{A}=\mathrm{sim}(\mathbf{f}_{1,b}^{Ad},\mathbf{f}_{1,k}^{As})/\tau_c$, and define the adapter-level contrastive objective as $\mathcal{L}_{1}^{A}=\mathcal{J}(\mathbf{S}^{A})$. To further guide the adapter toward cross-view compensation, we introduce a view-gap closure loss. Let $\bar{\mathbf{f}}_{1,b}^{d}$ and $\bar{\mathbf{f}}_{1,b}^{s}$ denote the normalized backbone features, and let $\bar{\mathbf{f}}_{1,b}^{Ad}$ and $\bar{\mathbf{f}}_{1,b}^{As}$ denote the normalized adapted features. The adapter-induced residual gap and the desired compensation direction are defined as
\begin{equation}
\mathbf{r}_{b}=
(\bar{\mathbf{f}}_{1,b}^{Ad}-\bar{\mathbf{f}}_{1,b}^{d})
-
(\bar{\mathbf{f}}_{1,b}^{As}-\bar{\mathbf{f}}_{1,b}^{s}),
\mathbf{c}_{b}=-\eta
(\bar{\mathbf{f}}_{1,b}^{d}-\bar{\mathbf{f}}_{1,b}^{s}).
\label{eq:view_gap_terms}
\end{equation}
Here, $\eta$ controls the compensation strength. $\mathbf{r}_{b}$ measures the relative correction introduced by the adapter between the two views, while $\mathbf{c}_{b}$ encourages this correction to close the original backbone-level view gap. Let $\mathbf{e}_{b}=\mathbf{r}_{b}-\mathbf{c}_{b}$. The view-gap closure loss is defined as
\begin{equation}
\mathcal{L}_{1}^{vg}
=
\frac{1}{B}
\sum_{b=1}^{B}
\left[
\mathcal{H}_{\delta}(\mathbf{e}_{b})
+
\frac{1}{4}
\big(
1-\cos(\mathbf{r}_{b},\mathbf{c}_{b})
\big)
\right],
\label{eq:view_gap_loss}
\end{equation}
where $\mathcal{H}_{\delta}(\cdot)$ denotes the element-wise averaged Huber penalty for robust compensation matching, and the cosine term is computed with stabilized normalization. The final Phase-II objective is
\begin{equation}
\mathcal{L}_{1}^{\mathrm{II}}
=
\mathcal{L}_{1}^{A}
+
\lambda_{\mathrm{1}}
\mathcal{L}_{1}^{vg},
\quad
t=1
\label{eq:cold_start_objective}
\end{equation}
where $\lambda_1$ is a weighting coefficient. Through this two-phase design, the backbone first learns a reliable cross-view embedding space, and the adapter then learns view-gap-aware residual corrections on top of this space. This provides a stable initialization for subsequent continual stages.

\subsection{Margin-Field Distillation}
\label{subsec:mfd}
For continual stages $t>1$, the model should preserve the cross-view geometry learned from previous environments to avoid catastrophic forgetting. We freeze the model from stage $t-1$ as a teacher and distill a compact \emph{margin field} on samples from $\mathcal{D}_t$. Unlike full affinity distillation, the margin field focuses on the margins between positive pairs and hard negatives, imposing selective geometric constraints around the retrieval boundary while retaining the plasticity needed for new environment adaptation. Specifically, given a mini-batch of $B$ matched cross-view pairs from $\mathcal{D}_t$, we compute the current stage student affinity as
\begin{equation}
(\mathbf{R}_{t})_{ij}
=
\beta_t
(\bar{\mathbf{f}}_{t,i}^{Ad})^{\top}
\bar{\mathbf{f}}_{t,j}^{As}.
\label{eq:mfd_affinity}
\end{equation}
Here, $\bar{\mathbf{f}}_{t,i}^{Ad}$ and $\bar{\mathbf{f}}_{t,j}^{As}$ denote the normalized adapted features of the current stage student. The frozen teacher affinity $(\mathbf{R}_{t-1})_{ij}$ is obtained analogously using the normalized adapted features produced by the stage $(t-1)$ teacher and its logit scale $\beta_{t-1}$. For the $i$-th query, the diagonal entry $(\mathbf{R}_{t-1})_{ii}$ corresponds to its positive match. We select the top-$K$ hard negatives from the off-diagonal entries of the teacher affinity matrix:
\begin{equation}
\mathcal{H}_{t-1,i}
=
\mathop{\mathrm{arg\,topK}}\limits_{j\neq i}
(\mathbf{R}_{t-1})_{ij},
\quad
|\mathcal{H}_{t-1,i}|=K.
\label{eq:hard_negative_set}
\end{equation}

For each selected negative $j\in\mathcal{H}_{t-1,i}$, the teacher and student margins are defined as $m_{t-1,i,j}=[(\mathbf{R}_{t-1})_{ii}-(\mathbf{R}_{t-1})_{ij}]/\tau_m$ and $m_{t,i,j}=[(\mathbf{R}_{t})_{ii}-(\mathbf{R}_{t})_{ij}]/\tau_m$, where $\tau_m$ is the margin temperature. Matching these margins preserves how much the positive match is preferred over hard negatives, rather than forcing all pairwise similarities to remain unchanged. 

Not all teacher margins are equally reliable under domain shift. We therefore introduce confidence-aware weights. Let $\mathcal{A}_{i}=\{i\}\cup\mathcal{H}_{t-1,i}$ denote the restricted candidate set. The teacher distribution over this set is computed as $p_{i\ell}=\mathrm{softmax}_{\ell\in\mathcal{A}_{i}}((\mathbf{R}_{t-1})_{i\ell}/\tau_d)$, and its entropy is $H_i=-\sum_{\ell\in\mathcal{A}_{i}}p_{i\ell}\log p_{i\ell}$. The row reliability is defined as $\omega_i=\mathrm{clip}(1-H_i/\log(K+1),\omega_{\min},1)$, and the margin confidence is defined as $a_{ij}=\max(\sigma(m_{t-1,i,j}),\omega_{\min})$. We combine them as $w_{ij}=\omega_i a_{ij}$. Let $e_{ij}=m_{t,i,j}-m_{t-1,i,j}$ and $Z=\sum_{i=1}^{B}\sum_{j\in\mathcal{H}_{t-1,i}}w_{ij}+\epsilon$. The one-directional margin-field distillation loss is written as
\begin{equation}
\mathcal{D}_{\mathrm{MFD}}(\mathbf{R}_{t},\mathbf{R}_{t-1})
=
\frac{1}{Z}
\sum_{i=1}^{B}
\sum_{j\in\mathcal{H}_{t-1,i}}
w_{ij}
\mathcal{H}_{\delta}(e_{ij}),
\label{eq:mfd_loss_one_direction}
\end{equation}
where $\mathcal{H}_{\delta}(\cdot)$ denotes the Huber penalty, and $\epsilon$ is a small constant for numerical stability. Since cross-view retrieval is bidirectional, we apply the same constraint to the transposed affinity matrices:
\begin{equation}
\mathcal{L}_{\mathrm{MFD}}
=
\frac{
\mathcal{D}_{\mathrm{MFD}}(\mathbf{R}_{t},\mathbf{R}_{t-1})
+
\mathcal{D}_{\mathrm{MFD}}(\mathbf{R}_{t}^{\top},\mathbf{R}_{t-1}^{\top})
}{2}.
\label{eq:mfd_symmetric}
\end{equation}

For $t>1$, For \(t>1\), the contrastive loss \(\mathcal{L}_{t}^{A}\) and the view-gap closure loss \(\mathcal{L}_{t}^{vg}\) are computed on the current stage data \(\mathcal{D}_{t}\) using the same definitions as at \(t=1\). We maintain plasticity with these two objectives
and mitigate forgetting through margin-field distillation. The
final objective is
\begin{equation}
\mathcal{L}_{t}
=
\underbrace{
\mathcal{L}^A_{t}
+
\lambda_{\mathrm{1}}\mathcal{L}_{t}^{vg}
}_{\text{Plasticity}}
+
\underbrace{
\lambda_{\mathrm{2}}\mathcal{L}_{\mathrm{MFD}}
}_{\text{Anti-forgetting}},
\quad t>1.
\label{eq:continual_objective}
\end{equation}
where $\lambda_1$ and $\lambda_2$ are weighting coefficients to control the relative importance of the loss terms.

\section{Experiments}
\subsection{Experimental Settings}

\textbf{Benchmarks}. 
We construct a continual DVGL evaluation setting from five representative datasets, including IR-VL328 \cite{liu2025object}, DenseUAV \cite{DenseUAV}, SUES-200 \cite{zhu2023sues}, CVGL-RGBT (C-RGBT) \cite{zhou2025cdm}, and University-1652 (U-1652) \cite{zheng2020university}, covering diverse scenes, viewpoints, platforms, and both visible and infrared modalities. For clarity, we denote them as $\mathcal{D}_1$ to $\mathcal{D}_5$ in the above order, and define $\mathcal{O}_0 = [\mathcal{D}_1,\mathcal{D}_2,\mathcal{D}_3,\mathcal{D}_4,\mathcal{D}_5]$ as the default dataset order. Unless otherwise specified, all experiments follow $\mathcal{O}_0$.

\noindent \textbf {Evaluation Metrics.} We evaluate retrieval accuracy with Recall@1 (R@1) and Average Precision (AP). Additionally, to summarize performance across continual stages, we report the Average (Avg) R@1 and AP over the five datasets. Knowledge retention on previously learned datasets is further evaluated using Forgetting (Fgt.) and Backward Transfer (BWT), both computed from R@1 following these works \cite{chaudhry2018riemannian,lin2022beyond}. Storage is also reported to quantify the model storage cost in MB.

\noindent\textbf{Implementation Details.} For CBS, both the backbone warm-up phase and the adapter activation phase are trained for $5$ epochs. We set $\lambda_1=0.10$ in Eq.~\eqref{eq:cold_start_objective} and Eq.~\eqref{eq:continual_objective}. For the Geo-Adapter, the residual scale $\alpha_t$ is initialized to $0.03$, and the adapter learning rate multiplier is set to $5$. The compensation ratio in the view-gap closure loss is set to $\eta=0.75$. For MFD, we select the top-$K=32$ hard negatives and set $\tau_m=1.0$, $\tau_d=4.0$, $\omega_{\min}=0.1$, and $\lambda_2=0.3$. Each continual stage after $t=1$ is trained for $5$ epochs. Further details are provided in the supplementary material.

\subsection{Comparison with State-of-the-Art Methods}

GeoMFD is compared with state-of-the-art DVGL methods, including Sample4Geo \cite{deuser2023sample4geo}, MEAN \cite{chen2024multi}, MFRGN \cite{wang2024mfrgn}, CAMP \cite{wu2024camp}, DAC \cite{xia2024enhancing}, and SURFNet \cite{liu2026surfnet}, as well as continual retrieval methods, including DKP \cite{xu2024distribution}, LSTKC++ \cite{11010188}, and DASK \cite{xu2025dask}. 

\noindent\textbf{Overall Comparison.}
As shown in Table~\ref{tab1}, GeoMFD achieves the best average performance in both retrieval directions under the single-model continual setting. It obtains R@1/AP scores of 89.44\%/87.48\% for Drone$\rightarrow$Satellite retrieval, exceeding the best individually trained method by 1.52\%/0.35\%. For Satellite$\rightarrow$Drone retrieval, it reaches 79.40\%/77.14\%, with improvements of 2.90\%/3.84\%. Compared with the strongest continual retrieval method, GeoMFD improves R@1/AP by 47.15\%/42.14\% and 28.41\%/38.85\% in the two retrieval directions, respectively. It also requires only 345.9\,MB of storage, substantially less than maintaining separate models for all datasets. These results demonstrate that GeoMFD achieves stronger overall performance than individually trained DVGL methods and existing continual retrieval methods, while maintaining a low storage cost.

\noindent\textbf{Sequential Fine-Tuning of DVGL Methods.}
As shown in Fig.~\ref{fig2}, we compare GeoMFD with existing DVGL methods under the sequential fine-tuning order $\mathcal{O}_0$. Compared with their independently trained results in Table~\ref{tab1}, conventional DVGL methods exhibit substantial performance degradation and large variations across datasets, particularly on the earlier environments. In contrast, GeoMFD maintains consistently strong performance across all five datasets and achieves the highest average R@1 and AP in both retrieval directions. Additionally, the zoomed-in regions on the latest dataset $\mathcal{D}_5$ show that this stability does not compromise adaptation to new environments. These results demonstrate that GeoMFD achieves a favorable balance between stability and plasticity, providing stronger resistance to catastrophic forgetting than direct sequential fine-tuning.

\begin{table}[t]
\centering
\small
\setlength{\tabcolsep}{2.1pt}
\begin{tabular}{l|cc|cc}
\hline
\multirow{2}{*}{Method}
& \multicolumn{2}{c|}{\textbf{Drone $\rightarrow$ Satellite}}
& \multicolumn{2}{c}{\textbf{Satellite $\rightarrow$ Drone}} \\
& R@1 & AP & R@1 & AP \\
\hline
Baseline & 81.05 & 79.91 & 74.83 & 70.72 \\
w/o CBS & 87.26 & 85.84 & 76.38 & 75.70 \\
w/o Geo-Adapter    & 88.54 & 86.73 & 78.23 & 76.45 \\
w/o MFD         & 82.38 & 81.24 & 75.50 &71.81 \\\hline
GeoMFD     & \textbf{89.44} & \textbf{87.48} & \textbf{79.40} & \textbf{77.14} \\
\hline
\end{tabular}
\caption{\textbf{Ablation study of GeoMFD.} 
Each component is removed from GeoMFD to evaluate its contribution, with average performance reported over the five datasets.}
\label{tab:ablation}
\end{table}

\begin{table}[t]
\centering
\small
\setlength{\tabcolsep}{1.2pt}
\begin{tabular}{c|cccc|cccc}
\hline
\multirow{2}{*}{\shortstack{After\\Stage $t$}}
& \multicolumn{4}{c|}{w/o MFD}
& \multicolumn{4}{c}{GeoMFD} \\
& R@1 & AP & Fgt.$\downarrow$ & BWT$\uparrow$
& R@1 & AP & Fgt.$\downarrow$ & BWT$\uparrow$ \\
\hline

\multicolumn{9}{c}{\textbf{Drone $\rightarrow$ Satellite}} \\
\hline
$t{=}3$ & 76.79 & 73.50 & 13.90 & -12.38
        & 84.00 & 79.40 & 2.51 & -1.08 \\
$t{=}4$ & 86.82 & 83.62 & 2.51 & -1.50
        & 87.77 & 85.08 & 0.79 & 0.52 \\
$t{=}5$ & 82.38 & 81.24 & 9.75 & -8.99
        & \textbf{89.44} & \textbf{87.48}
        & \textbf{0.93} & \textbf{0.36} \\
\hline

\multicolumn{9}{c}{\textbf{Satellite $\rightarrow$ Drone}} \\
\hline
$t{=}3$ & 64.68 & 61.66 & 7.55 & -5.60
        & 67.68 & 65.03 & 3.92 & 1.15 \\
$t{=}4$ & 75.48 & 71.36 & 0.63 & 0.66
        & 76.58 & 72.97 & 0.06 & 3.32 \\
$t{=}5$ & 75.50 & 71.81 & 5.93 & -4.96
        & \textbf{79.40} & \textbf{77.14}
        & \textbf{1.70} & \textbf{0.98} \\
\hline
\end{tabular}%
\caption{Stage-wise forgetting analysis of MFD. Results are averaged over all datasets seen up to stage $t$.}
\label{tab:mfd_stage_analysis}
\end{table}

\begin{figure*}[ht]
  \centering
  \includegraphics[width=6.9in]{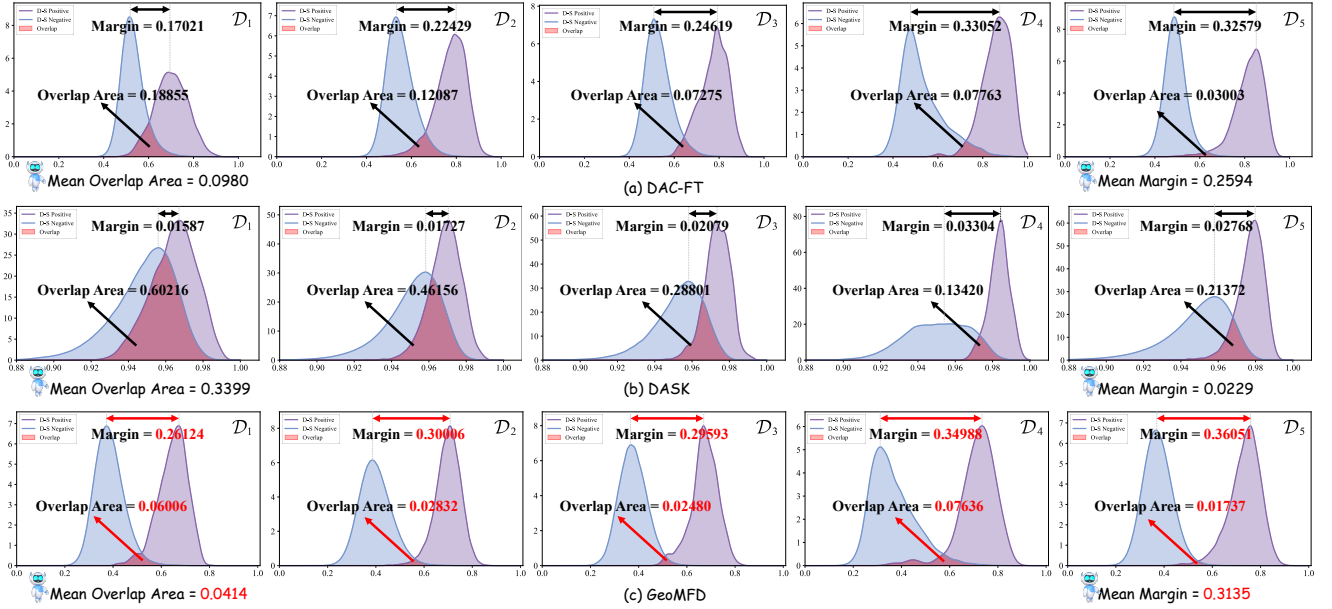}
    \caption{Visualization of cross-view similarity distributions for DAC-FT, DASK, and GeoMFD across $\mathcal{D}_{1}$--$\mathcal{D}_{5}$. The margin measures the separation between positive and negative pairs, and the overlap area indicates matching ambiguity. Larger margins and smaller overlap areas indicate better cross-view discrimination.}
  \label{fig3}
\end{figure*}

\subsection{Ablation Analysis}

\noindent\textbf{Component Ablation.}
As shown in Table~\ref{tab:ablation}, each component contributes consistently to the overall performance of GeoMFD. Removing MFD causes the most pronounced degradation, reducing the average R@1/AP from 89.44\%/87.48\% to 82.38\%/81.24\% for Drone$\rightarrow$Satellite retrieval and from 79.40\%/77.14\% to 75.50\%/71.81\% for Satellite$\rightarrow$Drone retrieval. This substantial drop highlights the importance of preserving historical similarity margins during sequential adaptation. CBS also provides clear benefits, particularly improving Satellite$\rightarrow$Drone R@1 by 3.02\%, which indicates that reliable initialization is important for subsequent continual learning. In addition, removing the Geo-Adapter decreases R@1 by 0.90\% and 1.17\% in the two retrieval directions, respectively, confirming the effectiveness of geometry-aware feature adaptation.

\noindent\textbf{Stage-Wise Forgetting Analysis.}
As shown in Table~\ref{tab:mfd_stage_analysis}, MFD consistently improves knowledge retention during sequential adaptation. At $t=3$, removing MFD results in Fgt./BWT values of 13.90\%/$-12.38$\% for Drone$\rightarrow$Satellite retrieval and 7.55\%/$-5.60$\% for Satellite$\rightarrow$Drone retrieval. With MFD, these values improve to 2.51\%/$-1.08$\% and 3.92\%/1.15\%, respectively. After all five stages, MFD further reduces Fgt. from 9.75\% to 0.93\% and from 5.93\% to 1.70\% in the two directions, and improves BWT from $-8.99$\% to 0.36\% and from $-4.96$\% to 0.98\%. These results demonstrate that MFD effectively mitigates catastrophic forgetting and promotes positive backward transfer by preserving historical cross-view similarity margins.
\begin{table}[t]
\centering
\small
\setlength{\tabcolsep}{2.5pt}
\begin{tabular}{c|cc|cc}
\hline
\multirow{2}{*}{Sequence}
& \multicolumn{2}{c|}{\textbf{Drone $\rightarrow$ Satellite}}
& \multicolumn{2}{c}{\textbf{Satellite $\rightarrow$ Drone}} \\
& R@1 & AP & R@1 & AP \\
\hline
$[\mathcal{D}_5,\mathcal{D}_1,\mathcal{D}_2,\mathcal{D}_4,\mathcal{D}_3]$
& 88.83 & 86.96 & 79.80 & 76.62 \\
$[\mathcal{D}_4,\mathcal{D}_2,\mathcal{D}_3,\mathcal{D}_5,\mathcal{D}_1]$
& 88.67 & 86.72 & 80.12 & \textbf{77.85} \\
$[\mathcal{D}_3,\mathcal{D}_4,\mathcal{D}_2,\mathcal{D}_1,\mathcal{D}_5]$
& 87.26 & 85.70 & 79.66 & 76.13 \\
$[\mathcal{D}_2,\mathcal{D}_3,\mathcal{D}_1,\mathcal{D}_5,\mathcal{D}_4]$
& \textbf{89.68} & \textbf{88.04} & \textbf{81.33} & 76.24 \\
$[\mathcal{D}_1,\mathcal{D}_2,\mathcal{D}_3,\mathcal{D}_4,\mathcal{D}_5]$
& 89.44 & 87.48 & 79.40 & 77.14 \\
\hline
\end{tabular}
\caption{\textbf{Effect of dataset order on C-DVGL.}
Average R@1 (\%) and AP (\%) are reported under different dataset orders.}
\label{tab:order_effect}
\end{table}
\subsection{Further Analysis}
\noindent\textbf{Robustness to Dataset Order.}
To evaluate the robustness of GeoMFD to different environment arrival orders,
Table~\ref{tab:order_effect} reports the results under five dataset
permutations. Across these orders, Drone$\rightarrow$Satellite R@1/AP ranges
from 87.26\%/85.70\% to 89.68\%/88.04\%, while
Satellite$\rightarrow$Drone R@1/AP ranges from 79.40\%/76.13\% to
81.33\%/77.85\%. The maximum variation is only 2.42\%, which demonstrates that
GeoMFD is robust to changes in dataset order. Moreover, no single order
consistently performs best across all metrics. These results confirm that GeoMFD remains robust across different dataset orders without relying on a favorable sequence.

\begin{table}[t]
\centering
\small
\setlength{\tabcolsep}{2.4pt}
\begin{tabular}{ll|cc|cc}
\hline
\multirow{2}{*}{Backbone}
& \multirow{2}{*}{Method}
& \multicolumn{2}{c|}{Drone $\rightarrow$ Satellite}
& \multicolumn{2}{c}{Satellite $\rightarrow$ Drone} \\
& & R@1 & AP & R@1 & AP \\
\hline

\multirow{2}{*}{ConvNeXt-S}
& Baseline
& 69.80 & 70.21 & 66.92 & 61.88 \\
& GeoMFD
& \textbf{74.61} & \textbf{74.53}
& \textbf{72.75} & \textbf{66.12} \\
\hline

\multirow{2}{*}{ConvNeXt-B}
& Baseline
& 73.64 & 73.58 & 71.95&64.80 \\
& GeoMFD
& \textbf{78.31} & \textbf{77.57}
& \textbf{75.01} & \textbf{68.50} \\
\hline

\multirow{2}{*}{ViT-S/16}
& Baseline
& 75.60 & 75.27 & 71.14 & 64.52 \\
& GeoMFD
& \textbf{84.06} & \textbf{82.73}
& \textbf{76.14} & \textbf{70.97} \\
\hline

\multirow{2}{*}{ViT-B/16}
& Baseline
& 81.05 & 79.91 & 74.83 & 70.72 \\
& GeoMFD
& \textbf{89.44} & \textbf{87.48}
& \textbf{79.40} & \textbf{77.14} \\
\hline
\multirow{2}{*}{ViT-L/16}
& Baseline
& 85.22 & 84.07 & 77.76 & 73.70 \\
& GeoMFD
& \textbf{91.62} & \textbf{89.70}
& \textbf{80.85} & \textbf{78.80} \\
\hline
\end{tabular}

\caption{GeoMFD is evaluated with ConvNeXt backbones and ViT backbones initialized with DINOv3 weights, and compared with the corresponding sequential fine-tuning baselines. Average results over five datasets are reported.}
\label{tab:backbone_generalization}
\end{table}

\noindent\textbf{Robustness across Backbones.}
To assess the robustness of GeoMFD across backbones, we conduct experiments with five backbones. As shown in Table~\ref{tab:backbone_generalization}, GeoMFD consistently outperforms sequential fine-tuning in both retrieval directions, with improvements of 3.06\%--8.46\% across all metrics. The results remain consistent for ConvNeXt backbones~\cite{liu2022convnet} and ViT backbones initialized with DINOv3. These results indicate that GeoMFD remains effective across different backbone families and model scales, rather than relying on a particular encoder capacity.

\noindent\textbf{Cross-View Similarity Distribution.}
To further examine the preservation of cross-view embedding geometry, Fig.~\ref{fig3} compares GeoMFD with DAC-FT and DASK. DAC-FT is the sequentially fine-tuned version of the DVGL method with the best overall independent-training performance, and DASK achieves the best overall performance among the continual retrieval methods. Across all five datasets, GeoMFD produces more clearly separated positive and negative similarity distributions, achieving the largest mean margin of 0.3135 and the smallest mean overlap area of 0.0414. Compared with DAC-FT, GeoMFD increases the mean margin from 0.2594 to 0.3135 and reduces the overlap area by 57.8\%. Compared with DASK, it reduces the overlap area from 0.3399 to 0.0414, corresponding to an 87.8\% reduction. These results confirm that GeoMFD more effectively preserves discriminative cross-view geometry and reduces ambiguity.

\section{Conclusion}
This paper investigates the practically important problem of continual drone-view geo-localization (C-DVGL), where a single model must adapt to sequentially arriving environments while retaining previously learned cross-view knowledge. We propose GeoMFD, a geometry-aware continual adaptation method that jointly supports environment adaptation and knowledge preservation. Extensive experiments on five DVGL datasets demonstrate that GeoMFD achieves superior overall performance compared with both independently trained DVGL methods and continual retrieval methods, while maintaining a single deployable model. This work provides a practical foundation for adaptive and reliable geo-localization in dynamic real-world environments.

\bibliography{aaai2027}


\clearpage

\twocolumn[
\begin{minipage}{\textwidth}
    \centering

    {\LARGE\bfseries
    GeoMFD: Continual Drone-View Geo-Localization with Geometry-Aware Adapter
and Margin-Field Distillation
    \par}

    \vspace{0.8em}

    {\Large\bfseries
    Supplementary Material
    \par}

    \vspace{1.8em}
\end{minipage}
]

\appendix

\subsection{Motivation and Distinction from Related Continual
Localization Settings}
\label{sec:supp_motivation}

\begin{figure*}[ht]
  \centering
  \includegraphics[width=6.5in]{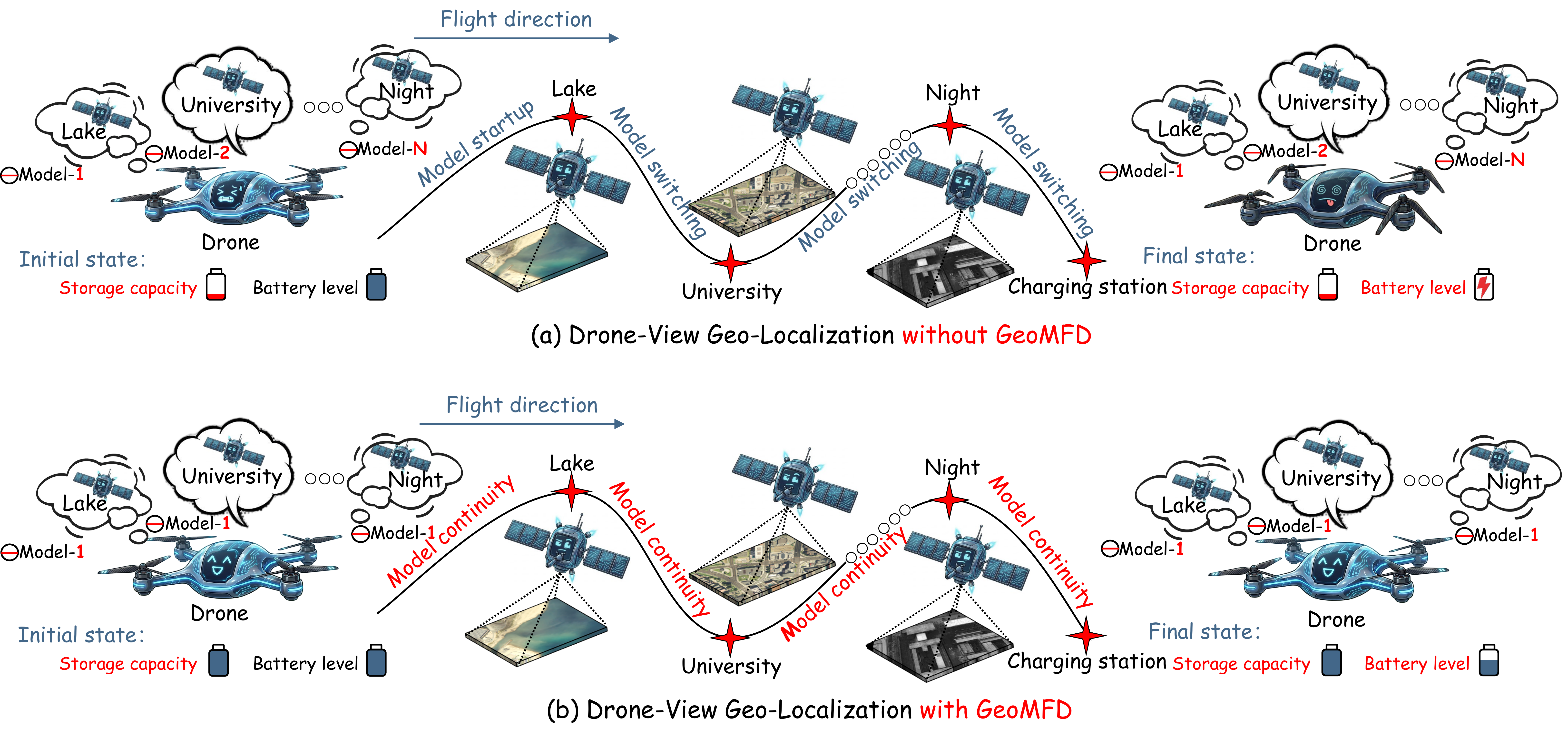}
  \caption{{Motivation of GeoMFD.} 
  (a) Conventional deployment may require multiple environment-specific models as the operating environment changes. This increases onboard storage, complicates deployment, and introduces additional model-selection overhead.
  (b) GeoMFD enables a single DVGL model to adapt across sequentially changing environments while maintaining continuity in the learned cross-view embedding geometry.
  }
  \label{fig4}
\end{figure*}

As shown in Fig.~\ref{fig4}, C-DVGL is motivated by the practical scenario
in which a drone successively operates in environments with different scene
structures, illumination conditions, and viewing characteristics. A static
DVGL model trained for a fixed environment may not remain effective as the
operating environment changes. Maintaining a separate model for each
environment increases onboard storage requirements and introduces additional
model-selection and switching overhead. In contrast, C-DVGL maintains a
single model that is sequentially updated across environments while retaining
the cross-view matching capability learned from previous ones. This requires
the model to preserve the relative relationships between positive pairs and
hard negatives during adaptation. The storage and battery symbols in
Fig.~\ref{fig4} conceptually illustrate onboard resource constraints rather
than measured hardware consumption.

Although continual learning has also been explored in visual
geo-localization~\cite{cheng2026first,shao2026towards}, the corresponding
continual scenarios differ from C-DVGL. HCL-Geo~\cite{cheng2026first}
constructs its task sequence according to changes in the field of view,
whereas the lifelong aerial VPR setting~\cite{shao2026towards} considers
sequential localization missions or domains. In contrast, C-DVGL focuses on
a DVGL model operating across sequentially changing drone environments.
Therefore, these settings differ mainly in how the continual localization
sequence is defined and in the deployment scenario being considered.

\section{Additional Method Details}
\label{sec:supp_method}

This section provides additional details of GeoMFD that complement the
formulation in the main paper. We first describe the group-wise spectral
calibration in Geo-Adapter and then analyze the geometric properties of its
reflection and tangent branches. We further clarify the motivation behind
margin-field distillation (MFD). We follow the notation introduced in the main
paper. Specifically, $\mathbf{f_t}\in\mathbb{R}^{C}$ denotes a backbone feature,
$\mathbf{\bar f_t}=\mathbf{f_t}/\lVert \mathbf{f_t}\rVert_2$ denotes its normalized representation, and
$\mathbf{f_t^{A}}=\bm{\Theta}_{t}^{A}(\mathbf{f}_{t})$ denotes the adapted feature. The same operations
are applied to both the drone-view and satellite-view branches.

\subsection{Backbone Architecture}
\label{sec:supp_backbone}

We adopt ViT-B/16 initialized with DINOv3 weights as the backbone for both the drone-view and
satellite-view branches. The input image is divided into non-overlapping
$16\times16$ patches and encoded by the Vision Transformer into a
$C=768$-dimensional feature representation. The two view branches use the
same backbone architecture and share their model parameters, thereby mapping
drone and satellite images into a common embedding space. The resulting
backbone feature $\mathbf{f}_t$ is subsequently processed by Geo-Adapter to
produce the environment-adapted representation $\mathbf{f}_t^{A}$.

\subsection{Group-Wise Spectral Calibration}
\label{sec:supp_spectral}

The spectral calibration extracts a low-dimensional descriptor of the
group-wise channel response, which is subsequently used to condition the
reflection and tangent residual branches. Given the backbone feature
$\mathbf{f}_t\in\mathbb{R}^{C}$, we first apply layer normalization:
\begin{equation}
    z_t=\operatorname{LN}(\mathbf{f}_t)\in\mathbb{R}^{C}.
    \label{eq:layer_norm_feature}
\end{equation}
We then divide the $C$ channels into $G$ non-overlapping groups, each
containing $P=C/G$ channels. Let
$\mathcal{I}_g=\{(g-1)P+1,\ldots,gP\}$ denote the channel indices of the
$g$-th group. Its response magnitude is summarized by the mean absolute
activation:
\begin{equation}
    (s_t)_g
    =
    \frac{1}{P}
    \sum_{c\in\mathcal{I}_g}
    \left|(z_t)_c\right|,
    \quad g=1,\ldots,G,
    \label{eq:spectral_stat}
\end{equation}
where $s_t\in\mathbb{R}^{G}$ provides a compact description of the
group-wise channel-response distribution.
In practice, Eq.~\eqref{eq:spectral_stat} is efficiently implemented by
adaptive average pooling over the absolute channel responses. The resulting
descriptor $s_t$ is passed through a lightweight bottleneck to predict a
bounded group-wise calibration vector:
\begin{equation}
    a_t = 
    \tanh\!\left(
    W_{s,2}\phi(W_{s,1}s_t+b_{s,1})+b_{s,2}
    \right),
    \label{eq:spectral_gate}
\end{equation}
where $a_t \in[-1,1]^G$. $\phi(\cdot)$ denotes the SiLU activation. Each entry of $\mathbf{a}_t$ is then broadcast to all channels within the corresponding group through $\mathcal{B}(\cdot)$, yielding the calibrated feature
\begin{equation}
    \mathbf{\widetilde f}_t
    =
    z_t\odot
    \left(
    \mathbf{1}_C+\mathcal{B}(a_t)
    \right),
    \label{eq:spectral_feature}
\end{equation}
where $\mathbf{1}_C$ is an all-one vector of dimension $C$. Thus, each channel
group is adaptively rescaled by a factor in $[0,2]$.

The use of absolute responses in Eq.~\eqref{eq:spectral_stat} allows $s_t$ to
capture the activation strength of each group independently of the response
sign. Meanwhile, the original signed feature $z_t$ is retained in
Eq.~\eqref{eq:spectral_feature}. The calibration therefore modulates the
relative importance of channel groups without discarding their signed response
patterns. In all experiments, we use $G=12$ groups for the $C=768$
ViT-B/16 embedding, and set the bottleneck width to $256$.

The weights and bias of the final linear layer in
Eq.~\eqref{eq:spectral_gate} are initialized to zero. Consequently,
$a_t=\mathbf{0}_G$ and $\widetilde f_t=z_t$ at initialization. This identity
initialization prevents the calibration module from perturbing the backbone
representation at the beginning of training, while allowing it to
progressively learn environment-conditioned channel reweighting.

\subsection{Geometric Properties of the Reflection and Tangent Residuals}
\label{sec:supp_geometry}

Given the calibrated feature $\widetilde{\mathbf{f}}_t$, Geo-Adapter predicts
two complementary geometric directions:
\begin{equation}
    \mathbf{v}_t
=
h_t^r(\widetilde{\mathbf{f}}_t)
\,/\,
\left\|
h_t^r(\widetilde{\mathbf{f}}_t)
\right\|_2,
    \quad
    \mathbf{q}_t
    =
    h_t^g(\widetilde{\mathbf{f}}_t),
    \label{eq:direction_prediction_supp}
\end{equation}
These definitions correspond to Eq.~(1) in the main paper. Here,
$\widetilde{\mathbf{f}}_t$ is the calibrated representation derived from the
original backbone feature $\mathbf{f}_t$ and is used to generate
input-dependent geometric directions. In contrast, the normalized feature
$\bar{\mathbf{f}}_t$ provides the geometric reference on the unit
hypersphere. The reflection residual is defined as
\begin{equation}
    \Delta_t^r
    =
    \left(
        \bar{\mathbf{f}}_t
        -
        2
        \left\langle
            \bar{\mathbf{f}}_t,\mathbf{v}_t
        \right\rangle
        \mathbf{v}_t
    \right)
    -
    \bar{\mathbf{f}}_t.
    \label{eq:reflection_supp}
\end{equation}
The transformed feature before residual subtraction can be written as
\begin{equation}
    \mathbf{f}_t^{\mathrm{ref}}
    =
    \bar{\mathbf{f}}_t
    -
    2
    \left\langle
        \bar{\mathbf{f}}_t,\mathbf{v}_t
    \right\rangle
    \mathbf{v}_t.
    \label{eq:reflected_feature_supp}
\end{equation}
Since
$\lVert\bar{\mathbf{f}}_t\rVert_2
=\lVert\mathbf{v}_t\rVert_2=1$,
the transformation applied to each input feature is a Householder reflection
and preserves its individual feature norm:
\begin{equation}
    \left\lVert
        \mathbf{f}_t^{\mathrm{ref}}
    \right\rVert_2^2
    =
    \left\lVert
        \bar{\mathbf{f}}_t
    \right\rVert_2^2
    -
    4
    \left\langle
        \bar{\mathbf{f}}_t,\mathbf{v}_t
    \right\rangle^2
    +
    4
    \left\langle
        \bar{\mathbf{f}}_t,\mathbf{v}_t
    \right\rangle^2
    \left\lVert
        \mathbf{v}_t
    \right\rVert_2^2
    =1.
    \label{eq:reflection_proof}
\end{equation}
Therefore, for each input feature, the reflection branch changes its angular
position on the unit hypersphere without altering its norm. Moreover,
$
    \left\lVert
        \Delta_t^r
    \right\rVert_2
    =
    2
    \left|
        \left\langle
            \bar{\mathbf{f}}_t,\mathbf{v}_t
        \right\rangle
    \right|
    \leq 2
$
shows that the reflection residual provides a bounded reflection residual, whose Euclidean norm corresponds to the chordal displacement between the original and reflected features
on the unit hypersphere. Additionally, the tangent residual is
defined as
\begin{equation}
    \Delta_t^g
    =
    \mathbf{q}_t
    -
    \left\langle
        \mathbf{q}_t,\bar{\mathbf{f}}_t
    \right\rangle
    \bar{\mathbf{f}}_t.
    \label{eq:tangent_supp}
\end{equation}
It removes the radial component of $\mathbf{q}_t$ with respect to
$\bar{\mathbf{f}}_t$. By construction,
\begin{equation}
    \left\langle
        \Delta_t^g,\bar{\mathbf{f}}_t
    \right\rangle
    =
    \left\langle
        \mathbf{q}_t,\bar{\mathbf{f}}_t
    \right\rangle
    -
    \left\langle
        \mathbf{q}_t,\bar{\mathbf{f}}_t
    \right\rangle
    \left\lVert
        \bar{\mathbf{f}}_t
    \right\rVert_2^2
    =
    0.
    \label{eq:tangent_proof}
\end{equation}
Hence,
$\Delta_t^g\in
T_{\bar{\mathbf{f}}_t}\mathbb{S}^{C-1}$,
where
$T_{\bar{\mathbf{f}}_t}\mathbb{S}^{C-1}$
denotes the tangent space of the unit hypersphere at
$\bar{\mathbf{f}}_t$. The tangent branch therefore provides a local angular
correction without directly moving the feature along its radial direction.
Moreover, since the tangent projection is non-expansive,
$\lVert\Delta_t^g\rVert_2\leq\lVert\mathbf{q}_t\rVert_2$.
This property controls the tangent residual relative to its predicted
direction, without requiring a sample-independent upper bound on its
magnitude.

Following Eq.~(3) in the main paper, the two residuals are adaptively fused as
\begin{equation}
    \Delta_t^A(\mathbf{f}_t)
    =
    \rho_t^r\Delta_t^r
    +
    \rho_t^g\Delta_t^g,
    \label{eq:mixture_supp}
\end{equation}
where the input-dependent mixture weights are predicted by
\begin{equation}
    [\rho_t^r,\rho_t^g]
    =
    \operatorname{softmax}
    \left(
        \psi_t(\mathbf{s}_t)
    \right).
    \label{eq:mixture_weights_supp}
\end{equation}
Since
$\rho_t^r,\rho_t^g\geq0$ and
$\rho_t^r+\rho_t^g=1$,
$\Delta_t^A(\mathbf{f}_t)$ is a convex combination of the reflection-based
residual and the local tangent residual. The notation
$\Delta_t^A(\mathbf{f}_t)$ emphasizes that the residual correction is
conditioned on the original input feature through its calibrated
representation and the corresponding channel-response statistics. Meanwhile,
the geometric construction of both residual branches is defined with respect
to $\bar{\mathbf{f}}_t$ on the unit hypersphere. The softmax gate therefore
adaptively balances the relative contributions of the two geometric
corrections according to the channel-response statistics of each input,
rather than imposing a uniform bound on the magnitude of the fused residual.
The overall correction strength is further modulated by the learnable
residual gate $\alpha_t$. The adapted feature is finally obtained as
\begin{equation}
    \mathbf{f}_t^A
    =
    \operatorname{Norm}
    \left(
        \bar{\mathbf{f}}_t
        +
        \alpha_t
        \Delta_t^A(\mathbf{f}_t)
    \right).
    \label{eq:adapter_output_supp}
\end{equation}
The final normalization maps the residual-updated feature back onto the unit
hypersphere, ensuring
$\lVert\mathbf{f}_t^A\rVert_2=1$.
When $\alpha_t=0$ or
$\Delta_t^A(\mathbf{f}_t)=\mathbf{0}$,
the normalization acts as an identity mapping on the already normalized
feature, yielding
$\mathbf{f}_t^A=\bar{\mathbf{f}}_t$.
For a sufficiently small $\alpha_t$, a first-order expansion around $\alpha_t=0$ gives
\begin{equation}
    \mathbf{f}_t^A
    =
    \bar{\mathbf{f}}_t
    +
    \alpha_t
    \left(
        \mathbf{I}
        -
        \bar{\mathbf{f}}_t
        \bar{\mathbf{f}}_t^{\top}
    \right)
    \Delta_t^A(\mathbf{f}_t)
    +
    \mathcal{O}(\alpha_t^2).
    \label{eq:adapter_first_order}
\end{equation}
This expression shows that, to first order, the final normalization removes
the radial component of $\Delta_t^A(\mathbf{f}_t)$, and only its component in
the tangent space at $\bar{\mathbf{f}}_t$ changes the embedding direction.
Since the expansion is performed around the unit-normalized base feature
$\bar{\mathbf{f}}_t$, the first-order term does not contain the factor
$1/\lVert\mathbf{f}_t\rVert_2$. Therefore, $\alpha_t$ directly controls the
first-order angular correction without being additionally rescaled by the norm
of the original feature.

We initialize $\alpha_t$ to $0.03$. The final linear layer of
$\psi_t(\cdot)$ is zero-initialized, yielding
$\rho_t^r=\rho_t^g=1/2$ before training. Consequently, Geo-Adapter starts
close to the normalized backbone embedding without introducing an arbitrary
initial preference for either geometric branch.

\subsection{Interpretation of View-Gap Closure}
\label{sec:supp_vgc}

For a matched pair, define the normalized backbone-level view gap as
$\mathbf{d}_b=
\bar{\mathbf{f}}_{1,b}^{d}
-
\bar{\mathbf{f}}_{1,b}^{s}$
and the relative adapter-induced correction as
\begin{equation}
    \mathbf{r}_b
    =
    \left(
        \bar{\mathbf{f}}_{1,b}^{Ad}
        -
        \bar{\mathbf{f}}_{1,b}^{d}
    \right)
    -
    \left(
        \bar{\mathbf{f}}_{1,b}^{As}
        -
        \bar{\mathbf{f}}_{1,b}^{s}
    \right).
\end{equation}
Accordingly, the adapted cross-view gap can be written as
\begin{equation}
    \bar{\mathbf{f}}_{1,b}^{Ad}
    -
    \bar{\mathbf{f}}_{1,b}^{As}
    =
    \mathbf{d}_b+\mathbf{r}_b.
\end{equation}
The target compensation direction in Eq.~(6) of the main paper is
$\mathbf{c}_b=-\eta\mathbf{d}_b$. Under the idealized condition
$\mathbf{r}_b=\mathbf{c}_b$, the adapted cross-view gap becomes
\begin{equation}
    \mathbf{d}_b+\mathbf{r}_b
    =
    (1-\eta)\mathbf{d}_b.
    \label{eq:gap_contraction}
\end{equation}
Therefore, $0<\eta<1$ encourages a partial contraction of the current
backbone-level view gap rather than its complete cancellation. In particular,
$\eta=1$ would target a zero gap for each matched pair, whereas a smaller
$\eta$ retains part of the original displacement and avoids imposing an
overly restrictive pairwise alignment that may suppress useful
discriminative variation with respect to nearby negatives. We use
$\eta=0.75$, corresponding to a target gap of
$0.25\mathbf{d}_b$ when the compensation target is matched exactly.

This interpretation describes the intended behavior of the loss and does not
require the equality $\mathbf{r}_b=\mathbf{c}_b$ to be satisfied exactly
during optimization. Since the backbone and adapter are jointly updated in
Phase II, $\mathbf{d}_b$ denotes the current backbone-level gap at each
optimization step, and the view-gap closure loss continuously guides the
relative adapter correction toward its corresponding compensation target.

The Huber term penalizes the component-wise discrepancy
$\mathbf{r}_b-\mathbf{c}_b$ and therefore provides a robust constraint on the
correction magnitude and structure. The cosine term in Eq.~(7) complements it
by explicitly encouraging directional consistency between
$\mathbf{r}_b$ and $\mathbf{c}_b$ while being insensitive to their absolute
scales. Thus, the two terms jointly constrain the magnitude and direction of
the adapter-induced compensation. Unlike the adapter-level contrastive
objective, which establishes batch-wise discrimination using positive and
negative pairs, the view-gap closure loss directly regularizes how the
adapter modifies the relative displacement within each matched cross-view
pair.

\subsection{Why Margin-Field Distillation Encourages Retrieval Geometry Preservation}
\label{sec:supp_mfd}

For a query \(i\) and a hard negative \(j\) selected by the teacher, we define the teacher affinity margin as
\begin{equation}
    \delta_{t-1,i,j}
    =
    (R_{t-1})_{ii}
    -
    (R_{t-1})_{ij}.
\end{equation}
Its sign determines the local retrieval order: the positive is ranked above the negative $j$ if and only if $\delta_{t-1,i,j}>0$. Matching the normalized
student margin
$m_{t,i,j}=\delta_{t,i,j}/\tau_m$
to the corresponding teacher margin
$m_{t-1,i,j}=\delta_{t-1,i,j}/\tau_m$
therefore preserves both the ranking decision and its separation from the
local decision boundary. In contrast to matching all $B^2$ affinities, margin
matching is invariant to a row-wise additive shift. For any scalar $c_i$,
\begin{equation}
    \big[(R_t)_{ii}+c_i\big]
    -
    \big[(R_t)_{ij}+c_i\big]
    =
    (R_t)_{ii}
    -
    (R_t)_{ij}.
    \label{eq:shift_invariance}
\end{equation}
MFD thus encourages the preservation of the selected local retrieval
geometry without requiring the student to reproduce domain-specific
row-wise affinity offsets.

In Eq.~(10) of the main paper, $K$ denotes the predefined maximum number of
hard negatives. Since each affinity row contains at most $B-1$
off-diagonal candidates, the effective number used in practice is
\begin{equation}
    K'
    =
    \min(K,B-1).
    \label{eq:effective_hard_negatives}
\end{equation}
Accordingly, for each query, the teacher selects
$|\mathcal{H}_{t-1,i}|=K'$ hard negatives, where $K=32$. The restricted
candidate set
$\mathcal{A}_i=\{i\}\cup\mathcal{H}_{t-1,i}$
therefore contains $K'+1$ elements. When $B-1\geq K$, we have $K'=K$, and
the formulation reduces exactly to that presented in the main paper. We use
mini-batches with $B\geq2$, ensuring that at least one in-batch negative is
available for each query.

Restricting distillation to these negatives concentrates the constraint near
the retrieval boundary and reduces the number of one-directional loss terms
from $O(B^2)$ to $O(BK')$. The affinity matrix is still computed in the
current implementation for in-batch retrieval and top-$K'$ selection, but no
historical image, feature, or affinity matrix is stored.

The confidence terms introduced in the main paper play complementary roles.
The normalized-entropy reliability is defined as
\begin{equation}
    \omega_i
    =
    \operatorname{clip}\!\left(
        1-\frac{H(p_i)}{\log(K'+1)},
        \omega_{\min},
        1
    \right).
\end{equation}
The denominator $\log(K'+1)$ is the maximum entropy of a distribution over
the positive and the $K'$ selected negatives. Thus, the normalization remains
valid when the available number of in-batch negatives is smaller than the
predefined maximum $K$. The reliability approaches one when the teacher
distribution over the positive and selected negatives is concentrated, and
approaches $\omega_{\min}$ when the distribution is ambiguous. The pairwise
margin confidence
\begin{equation}
    a_{ij}
    =
    \max\!\left(
        \sigma(m_{t-1,i,j}),
        \omega_{\min}
    \right)
\end{equation}
assigns greater weight to comparisons for which the teacher exhibits a clear
positive preference. Their product,
$w_{ij}=\omega_i a_{ij}$, prevents either an uncertain teacher row or an
unreliable pairwise relation from dominating the update, while the floor
$\omega_{\min}=0.1$ avoids discarding such supervision entirely. The Huber
penalty further limits the influence of large teacher--student discrepancies
caused by domain shift. Finally, MFD is applied in both retrieval directions
by performing the same construction on $R_t$ and $R_t^{\top}$.

\section{Details of the Benchmarks and Evaluation Metrics}
\label{sec:supp_datasets}

We conduct experiments on five drone-view geo-localization benchmarks with
distinct scene distributions, acquisition conditions, and cross-view gaps.
For all datasets, we evaluate
both retrieval directions. In Drone$\rightarrow$Satellite retrieval, drone
images are used as queries to retrieve their corresponding satellite images
from the gallery. Conversely, in Satellite$\rightarrow$Drone retrieval,
satellite images serve as queries to retrieve the matching drone images.

\paragraph{IR-VL328.}
IR-VL328 is an infrared--visible cross-modal benchmark designed for UAV
geo-localization under nighttime conditions \cite{liu2025object}. It contains $328$ geographically
independent scenes collected in real environments. The drone-view images are
captured at night using an onboard infrared camera, whereas the corresponding
reference images are obtained from visible-light satellite imagery. Compared
with conventional RGB-based benchmarks, IR-VL328 introduces not only a large
viewpoint discrepancy but also a substantial modality and illumination gap
between thermal aerial observations and visible satellite references. It
therefore provides a challenging environment for evaluating whether a
continual model can acquire cross-modal matching knowledge without disrupting
the visual correspondence learned from other environments.

\paragraph{DenseUAV.}
DenseUAV is a real-world benchmark developed for vision-based UAV
self-positioning in low-altitude urban environments  \cite{DenseUAV}. It contains more than
$27{,}000$ densely sampled drone-view and satellite-view images collected
across $14$ university campuses. The UAV images are acquired at flight
altitudes of approximately $80$, $90$, and $100$\,m, resulting in substantial
variations in image scale and ground coverage. Unlike datasets organized
around a small number of isolated landmarks, DenseUAV densely samples
spatially adjacent locations, producing many visually similar candidates in
the gallery. Repetitive building layouts, roads, vegetation, and neighboring
urban structures consequently form difficult hard negatives, making the
dataset particularly suitable for evaluating fine-grained retrieval geometry
and resistance to catastrophic forgetting.

\begin{table*}[ht]
\centering
\setlength{\tabcolsep}{3.0pt}
\renewcommand{\arraystretch}{1.05}
\begin{tabular}{l|l|cc|cc|cc|cc|cc|cc|c}
\hline
\multirow{2}{*}{Protocol}
& \multirow{2}{*}{Method}
& \multicolumn{2}{c|}{IR-VL328}
& \multicolumn{2}{c|}{DenseUAV}
& \multicolumn{2}{c|}{SUES-200}
& \multicolumn{2}{c|}{C-RGBT}
& \multicolumn{2}{c|}{U-1652}
& \multicolumn{2}{c|}{Avg.}
& \multirow{2}{*}{Storage} \\
& & R@1 & AP
& R@1 & AP
& R@1 & AP
& R@1 & AP
& R@1 & AP
& R@1 & AP
& \\
\hline

\multicolumn{15}{c}{\textbf{Drone $\rightarrow$ Satellite}} \\
\hline

\multirow{6}{*}{Individual}
& Sample4Geo
& 68.19 & 73.43
& 85.54 & 67.24
& 92.60 & 94.00
& 89.33 & 91.46
& 92.65 & 93.81
& 85.66 & 83.99
& 1752 \\

& MEAN
& 63.75 & 70.11
& 86.61 & 67.89
& 95.50 & 96.46
& 81.33 & 85.31
& 93.55 & 94.53
& 84.15 & 82.86
& 730.5 \\

& MFRGN
& 67.81 & 73.25
& 93.47 & 80.12
& 95.85 & 96.52
& 88.00 & 90.50
& 94.33 & 95.24
& 87.89 & 87.13
& 1869.5 \\

& CAMP
& 67.04 & 72.60
& 84.47 & 66.02
& 95.40 & 96.38
& 85.33 & 88.03
& 94.46 & 95.38
& 85.34 & 83.68
& 1828.5 \\

& DAC
& 71.15 & 75.56
& 85.67 & 68.03
& 96.80 & 97.54
& 91.32 & 93.12
& 94.67 & 95.50
& 87.92 & 85.95
& 1931 \\

& SURFNet
& 65.19 & 70.87
& 77.52 & 58.73
& 93.63 & 94.84
& 88.00 & 90.56
& 94.57 & 95.49
& 83.78 & 82.10
& 2124 \\
\hline

\multirow{8}{*}{Continual}
& DKP
& 22.42 & 29.10
& 20.93 & 11.41
& 48.64 & 56.19
& 40.74 & 48.10
& 58.87 & 63.72
& 38.32 & 41.70
& 409.0 \\

& LSTKC++
& 31.25 & 38.22
& 17.16 & 10.99
& 21.10 & 27.86
& 90.00 & 92.11
& 12.33 & 15.93
& 34.37 & 37.02
& 126.1 \\

& DASK
& 22.45 & 28.51
& 14.84 & 9.11
& 51.17 & 57.52
& 78.66 & 82.23
& 44.32 & 49.34
& 42.29 & 45.34
& 126.1 \\
\cline{2-15}

& GeoMFD ($\mathcal{O}_0$)
& 86.76 & 89.25
& 73.70 & 58.85
& 99.17 & 99.36
& 91.33 & 93.00
& 96.23 & 96.93
& 89.44 & 87.48
& 345.9 \\

& GeoMFD ($\mathcal{O}_1$)
& 85.23 & 88.22
& 71.47 & 56.64
& 98.72 & 98.90
& 94.00 & 95.35
& 94.71 & 95.69
& 88.83 & 86.96
& 345.9 \\

& GeoMFD ($\mathcal{O}_2$)
& 88.38 & 90.62
& 72.29 & 56.94
& 98.30 & 98.62
& 89.33 & 91.52
& 95.04 & 95.91
& 88.67 & 86.72
& 345.9 \\

& GeoMFD ($\mathcal{O}_3$)
& 84.18 & 87.20
& 68.21 & 54.31
& 99.33 & 99.46
& 88.67 & 90.90
& 95.93 & 96.64
& 87.26 & 85.70
& 345.9 \\

& GeoMFD ($\mathcal{O}_4$)
& 82.88 & 86.25
& 78.46 & 64.40
& 96.82 & 97.49
& 96.00 & 96.83
& 94.25 & 95.21
& 89.68 & 88.04
& 345.9 \\
\hline

\multicolumn{15}{c}{\textbf{Satellite $\rightarrow$ Drone}} \\
\hline

\multirow{6}{*}{Individual}
& Sample4Geo
& 75.78 & 63.92
& 20.45 & 18.54
& 97.50 & 93.63
& 88.00 & 90.65
& 95.14 & 91.39
& 75.37 & 71.63
& 1752 \\

& MEAN
& 68.75 & 57.36
& 20.33 & 18.26
& 97.50 & 94.75
& 86.66 & 89.45
& 96.01 & 92.08
& 73.85 & 70.38
& 730.5 \\

& MFRGN
& 78.12 & 59.33
& 23.35 & 21.40
& 98.75 & 94.42
& 84.00 & 87.48
& 96.15 & 93.94
& 76.07 & 71.31
& 1869.5 \\

& CAMP
& 71.87 & 63.37
& 19.56 & 17.71
& 96.25 & 93.36
& 83.33 & 86.42
& 96.15 & 92.72
& 73.43 & 70.72
& 1828.5 \\

& DAC
& 75.00 & 65.38
& 20.25 & 18.55
& 97.50 & 94.06
& 93.33 & 94.71
& 96.43 & 93.79
& 76.50 & 73.30
& 1931 \\

& SURFNet
& 71.88 & 59.01
& 17.61 & 16.07
& 95.00 & 90.75
& 89.33 & 91.13
& 95.72 & 93.20
& 73.91 & 70.03
& 2124 \\
\hline

\multirow{8}{*}{Continual}
& DKP
& 19.53 & 16.46
& 12.05 & 10.83
& 61.25 & 49.92
& 44.00 & 51.53
& 79.03 & 59.21
& 43.17 & 37.59
& 409.0 \\

& LSTKC++
& 71.09 & 22.41
& 13.94 & 12.26
& 36.25 & 17.53
& 91.33 & 92.94
& 42.36 & 10.85
& 50.99 & 31.20
& 126.1 \\

& DASK
& 25.00 & 10.35
& 6.58 & 6.33
& 52.50 & 42.39
& 84.00 & 87.30
& 77.31 & 45.07
& 49.08 & 38.29
& 126.1 \\
\cline{2-15}

& GeoMFD ($\mathcal{O}_0$)
& 90.62 & 81.35
& 17.68 & 16.63
& 100.00 & 98.28
& 92.00 & 93.84
& 96.71 & 95.59
& 79.40 & 77.14
& 345.9 \\

& GeoMFD ($\mathcal{O}_1$)
& 92.97 & 79.30
& 17.49 & 16.65
& 98.75 & 98.15
& 92.67 & 94.50
& 97.15 & 94.50
& 79.80 & 76.62
& 345.9 \\

& GeoMFD ($\mathcal{O}_2$)
& 94.53 & 85.88
& 17.13 & 16.45
& 98.75 & 97.40
& 93.33 & 95.00
& 96.86 & 94.51
& 80.12 & 77.85
& 345.9 \\

& GeoMFD ($\mathcal{O}_3$)
& 92.19 & 77.09
& 16.82 & 15.90
& 100.00 & 98.34
& 92.00 & 93.89
& 97.29 & 95.43
& 79.66 & 76.13
& 345.9 \\

& GeoMFD ($\mathcal{O}_4$)
& 98.43 & 77.86
& 19.05 & 18.07
& 97.50 & 95.70
& 94.66 & 96.00
& 97.00 & 93.56
& 81.33 & 76.24
& 345.9 \\
\hline
\end{tabular}%
\caption{Comparison with state-of-the-art methods under different
training protocols and continual dataset orders.
Individual methods are independently trained on each dataset, whereas
continual methods sequentially update a single model. GeoMFD is evaluated
under the five dataset orders defined in Eq.~\eqref{eq:continual_orders}.
For SUES-200, results are reported at 150\,m. Storage is measured in MB.}
\label{tab6}
\end{table*}

\paragraph{SUES-200.}
SUES-200 is a multi-height and multi-scene benchmark containing $200$
geographical locations and $24{,}120$ drone and satellite images  \cite{zhu2023sues}. The drone
images are captured at four flight altitudes, namely $150$, $200$, $250$, and
$300$\,m, resulting in systematic variations in apparent scale, visible
context, and spatial coverage. Its official split contains $120$ training
locations and $80$ test locations. The dataset covers diverse urban scenes and
is therefore suitable for evaluating robustness to altitude-induced scale
changes and scene-level appearance variations. Among the four altitude
settings, $150$\,m represents the most challenging retrieval condition because
the narrower ground coverage provides less contextual information for
cross-view matching. Accordingly, we report the $150$-m results in the main
paper, while the results at the remaining altitudes are provided in the
supplementary material.

\paragraph{C-RGBT.}
C-RGBT is a cross-modal cross-view geo-localization dataset composed of
paired visible and thermal images \cite{zhou2025cdm}. Its thermal images were
captured using a DJI H20T thermal module mounted on a DJI M300 RTK drone.
Following Urban-500, the dataset categorizes scenes into building-class and
nonbuilding-class scenarios, containing $503$ and $395$ RGBT image pairs,
respectively. The pronounced modality gap between visible and thermal imagery
provides a challenging setting for evaluating continual adaptation across
heterogeneous sensing conditions.

\paragraph{U-1652.}
U-1652 is a large-scale multi-view and multi-source benchmark
covering $1{,}652$ buildings from $72$ universities worldwide \cite{zheng2020university}. It provides
images from synthetic drone cameras, satellites, and ground cameras; only the
drone and satellite modalities are used in our experiments. Each target is
observed from multiple drone viewpoints around the corresponding building,
introducing pronounced changes in scale, orientation, and visible facade
content. The training set contains $701$ buildings from $33$ universities,
while the evaluation locations are drawn from $39$ disjoint universities.
The satellite gallery contains $951$ buildings, including $250$ additional
distractor locations, which increases the difficulty of large-scale
retrieval. Owing to its broad geographical coverage and diverse architectural
styles, U-1652 provides a comprehensive final environment for
evaluating both cross-view generalization and accumulated forgetting.

Unless otherwise specified, the default continual sequence is
IR-VL328 $\rightarrow$ DenseUAV $\rightarrow$ SUES-200 $\rightarrow$
C-RGBT $\rightarrow$ U-1652. At stage $t$, optimization uses only
images from the current dataset $\mathcal{D}_t$. The checkpoint obtained at
stage $t-1$ initializes the student model, while a frozen copy of the same
checkpoint serves as the teacher throughout stage $t$. We use neither exemplar replay nor any stored images, features, or affinity matrices from previous datasets. After each stage, the resulting single checkpoint is
evaluated on the test sets of all environments observed so far. Dataset
identities are not provided to the model during inference.

\subsection{Metrics and Storage Accounting}
\label{sec:supp_metrics}

Let $a_{t,k}$ denote R@1 on dataset $D_k$ after training stage $t$, where
$k\leq t$. The average accuracy after stage $t$ is
\begin{equation}
    A_t=\frac{1}{t}\sum_{k=1}^{t}a_{t,k}.
\end{equation}
Following the continual-learning convention used in the main paper, forgetting
and backward transfer are computed as
\begin{equation}
\operatorname{Fgt}_t=\frac{1}{t-1}\sum_{k=1}^{t-1}
\left(\max_{\ell\in\{k,\ldots,t-1\}}a_{\ell,k}-a_{t,k}\right),
\end{equation}
\begin{equation}
\operatorname{BWT}_t=\frac{1}{t-1}\sum_{k=1}^{t-1}(a_{t,k}-a_{k,k}),
\quad t>1.
\end{equation}
Lower forgetting and higher BWT indicate better retention; negative BWT
denotes performance loss on earlier environments. At the final stage, the Avg
values in the main tables are the arithmetic mean of R@1 or AP over all five
datasets, computed separately for Drone$\rightarrow$Satellite and
Satellite$\rightarrow$Drone retrieval.

Storage is measured as the total size, in MB, of parameters required for
deployment after the full sequence. For an independently trained method, this
includes one complete model per environment. For a continual method, it
includes only the final deployable model; the frozen teacher is a transient
training copy and is not retained for inference. Dataset files and temporary
optimizer states are excluded for all methods.

\begin{table*}[t]
\centering
\small
\setlength{\tabcolsep}{2.0pt}
\renewcommand{\arraystretch}{1.05}
\begin{tabular}{
l|l|
cccccccc|
cccccccc
}
\hline

\multicolumn{2}{c}{}
& \multicolumn{16}{c}{SUES-200} \\
\cline{1-18}

\multirow{3}{*}{Protocol}
& \multirow{3}{*}{Method}
& \multicolumn{8}{c|}{Drone$\rightarrow$Satellite}
& \multicolumn{8}{c}{Satellite$\rightarrow$Drone} \\
\cline{3-18}

&
& \multicolumn{2}{c}{150 m}
& \multicolumn{2}{c}{200 m}
& \multicolumn{2}{c}{250 m}
& \multicolumn{2}{c|}{300 m}
& \multicolumn{2}{c}{150 m}
& \multicolumn{2}{c}{200 m}
& \multicolumn{2}{c}{250 m}
& \multicolumn{2}{c}{300 m} \\
\cline{3-18}

&
& R@1 & AP
& R@1 & AP
& R@1 & AP
& R@1 & AP
& R@1 & AP
& R@1 & AP
& R@1 & AP
& R@1 & AP \\
\hline

\multirow{8}{*}{Individual}
& APA-BI
& 86.45 & 88.90
& 91.85 & 93.54  
& 95.70 & 96.62
& 96.70 & 97.37
& 97.50 & 86.21
& 98.75 & 94.03
& 98.75 & 96.95
& 98.75 & 97.82  \\

& SHAA
& 90.32 & 92.09
& 96.50 & 97.16
& 97.30 & 97.71
& 97.40 & 97.93
& 97.50 & 91.50
& 98.75 & 95.68
& 98.75 & 97.45
& 98.75 & 97.31 \\

& Sample4Geo
& 92.60 & 94.00
& 97.38 & 97.81
& 98.28 & 98.64
& 99.18 & 99.36
& 97.50 & 93.63
& 98.75 & 96.70
& 98.75 & 98.28
& 98.75 & 98.05 \\

& DAC
& 96.80 & 97.54
& 97.48 & 97.97
& 98.20 & 98.62
& 97.58 & 98.14
& 97.50 & 94.06
& 98.75 & 96.66
& 98.75 & 98.09
& 98.75 & 97.87 \\

& CAMP
& 95.40 & 96.38
& 97.63 & 98.16
& 98.05 & 98.45
& 99.33 & 99.46
& 96.25 & 93.69
& 97.50 & 96.76
& 98.75 & 98.10
& 100.00 & 98.85 \\

& CDM-Net
& 93.78 & 95.16
& 97.62 & 98.16
& 98.28 & 98.69
& 99.20 & 99.31
& 95.25 & 92.24
& 98.50 & 96.40
& 99.00 & 97.60
& 99.00 & 98.01 \\

& GeoBridge
& 95.68 & 97.49
& 97.88 & 98.89
& 97.85 & 98.84
& 97.63 & 98.62
& 98.99 & 97.98
& 98.98 & 97.98
& 99.01 & 97.98
& 98.75 & 97.95 \\

& MEAN
& 95.50 & 96.46
& 98.38 & 98.72
& 98.95 & 99.17
& 99.52 & 99.63
& 97.50 & 94.75
& 100.00 & 97.09
& 100.00 & 98.28
& 100.00 & 99.21 \\
\hline

\multirow{5}{*}{Continual}
& GeoMFD ($\mathcal{O}_0$)
& {99.17} &{99.36}
& 99.70 & 99.77
& 99.95 & 99.96
& 99.90 & 99.92
& {100.00} &{98.28}
& 100.00 & 98.89
& 98.75 & 98.96
& 98.75 & 98.96 \\

& GeoMFD ($\mathcal{O}_1$)
& 98.72 & 98.90
& 99.08 & 99.21
& 99.23 & 99.35
& 99.35 & 99.48
& 98.75 & 98.15
& 98.75 & 98.71
& 98.75 & 98.60
& 98.75 & 98.86 \\

& GeoMFD ($\mathcal{O}_2$)
& 98.30 & 98.62
& 99.35 & 99.51
& 99.60 & 99.69
& 99.82 & 99.87
& 98.75 & 97.40
& 100.00 & 98.53
& 100.00 & 98.95
& 98.75 & 98.89 \\

& GeoMFD ($\mathcal{O}_3$)
& 99.33 & 99.46
& 99.80 & 99.85
& 99.82 & 99.87
& 99.95 & 99.96
& 100.00 & 98.34
& 100.00 & 98.27
& 98.75 & 98.55
& 98.75 & 98.68 \\

& GeoMFD ($\mathcal{O}_4$)
& 96.82 & 97.49
& 98.25 & 98.57
& 98.40 & 98.73
& 98.83 & 99.11
& 97.50 & 95.70
& 98.75 & 97.95
& 98.75 & 98.37
& 98.75 & 98.77  \\
\hline

\end{tabular}%
\caption{Comparison with state-of-the-art methods at four flight altitudes on
SUES-200. Individual methods are trained exclusively on SUES-200, whereas
GeoMFD is evaluated after completing each continual sequence. SUES-200 is
introduced at the first stage under $\mathcal{O}_3$.}
\label{tab:sues_all_heights}
\end{table*}

\section{Additional Experiments Details}
\subsection{Experimental Settings}
\label{sec:supp_optimization}

The shared backbone is DINOv3-ViT-B/16 pretrained on its original pretraining
corpus. All input images are resized to $384\times384$. The embedding
dimension is $C=768$. We use AdamW with a backbone learning rate of
$10^{-5}$. The adapter learning rate is multiplied by 5, giving \(5\times10^{-5}\), and the adapter weight decay is set to zero. A cosine
learning-rate schedule is restarted at every continual stage, with linear
warm-up over the first $10\%$ of the stage iterations. Gradients are
value-clipped to $100$. Training uses automatic mixed precision.
For CBS, the backbone warm-up and adapter-activation phases are trained for
five epochs each, resulting in a total of ten epochs at the initial stage.
To ensure a fair comparison, all corresponding ablation variants are
allocated the same total optimization budget at this stage, regardless of
whether the two-phase schedule is used. Every subsequent continual stage is
trained for five epochs. All experiments are conducted on the Pytorch deep learning framework with the experimental platform running on Ubuntu 22.04 equipped with four NVIDIA RTX 4090 GPUs.

\subsection{Additional Results under Different Dataset Orders}
\label{sec:supp_order}

To examine whether GeoMFD depends on a particular sequence of environments,
we further evaluate it under five continual dataset orders. We denote
IR-VL328, DenseUAV, SUES-200, C-RGBT, and U-1652 as
$\mathcal{D}_1$, $\mathcal{D}_2$, $\mathcal{D}_3$,
$\mathcal{D}_4$, and $\mathcal{D}_5$, respectively. The evaluated orders
are defined as
\begin{equation}
\begin{aligned}
\mathcal{O}_0 &=
[\mathcal{D}_1,\mathcal{D}_2,\mathcal{D}_3,
 \mathcal{D}_4,\mathcal{D}_5],\\
\mathcal{O}_1 &=
[\mathcal{D}_5,\mathcal{D}_1,\mathcal{D}_2,
 \mathcal{D}_4,\mathcal{D}_3],\\
\mathcal{O}_2 &=
[\mathcal{D}_4,\mathcal{D}_2,\mathcal{D}_3,
 \mathcal{D}_5,\mathcal{D}_1],\\
\mathcal{O}_3 &=
[\mathcal{D}_3,\mathcal{D}_4,\mathcal{D}_2,
 \mathcal{D}_1,\mathcal{D}_5],\\
\mathcal{O}_4 &=
[\mathcal{D}_2,\mathcal{D}_3,\mathcal{D}_1,
 \mathcal{D}_5,\mathcal{D}_4].
\end{aligned}
\label{eq:continual_orders}
\end{equation}

The detailed results under different continual dataset orders are reported
in Table~\ref{tab6}. GeoMFD maintains consistently strong performance in both
retrieval directions. For Drone$\rightarrow$Satellite retrieval, the average
R@1 and AP vary within 87.26\%--89.68\% and 85.70\%--88.04\%,
respectively. For Satellite$\rightarrow$Drone retrieval, the corresponding
ranges are 79.40\%--81.33\% and 76.13\%--77.85\%. The limited variation
across the five orders indicates that the overall retrieval performance is
robust to changes in the continual dataset sequence.

The results on individual datasets also show that no single order consistently
performs best. For Drone$\rightarrow$Satellite retrieval, different orders
achieve the strongest results on IR-VL328, DenseUAV, SUES-200, C-RGBT, and
U-1652. A similar pattern is observed for Satellite$\rightarrow$Drone
retrieval, where the best R@1 and AP values are distributed across several
orders. In terms of average performance, $\mathcal{O}_4$ obtains the highest
Drone$\rightarrow$Satellite R@1/AP and Satellite$\rightarrow$Drone R@1,
whereas $\mathcal{O}_2$ achieves the highest Satellite$\rightarrow$Drone AP.
Thus, the relative advantage of each order varies across datasets, metrics,
and retrieval directions.

Notably, the default order $\mathcal{O}_0$ used in the main experiments is
not uniformly optimal, yet it remains competitive with the other orders.
These observations show that the performance of GeoMFD is not driven by a
carefully selected dataset sequence. Instead, it remains stable across
different continual orders in both retrieval directions, supporting its
robustness to variations in the arrival order of environments.

\begin{table}[t]
\centering
\small
\setlength{\tabcolsep}{2.6pt}
\renewcommand{\arraystretch}{1.05}
\begin{tabular}{l|l|cc|cc}
\hline
\multirow{2}{*}{Protocol}
& \multirow{2}{*}{Method}
& \multicolumn{2}{c|}{Drone$\rightarrow$Satellite}
& \multicolumn{2}{c}{Satellite$\rightarrow$Drone} \\
&
& R@1 & AP
& R@1 & AP \\
\hline

\multirow{10}{*}{Individual}
& APA-BI
& 93.57 & 94.55
& 95.86 & 92.88 \\

& SHAA
& 93.69 & 94.68
& 96.15 & 93.49 \\

& Sample4Geo
& 92.65 & 93.81
& 95.14 & 91.39 \\

& MEAN
& 93.55 & 94.53
& 96.01 & 92.08 \\

& MFRGN
& 94.33 & 95.24
& 96.15 & 93.94 \\

& CAMP
& 94.46 & 95.38
& 96.15 & 92.72 \\

& DAC
& 94.67 & 95.50
& 96.43 & 93.79 \\

& SURFNet
& 94.57 & 95.49
& 95.72 & 93.20 \\

& CDM-Net
& {95.13} &{96.04}
& 96.68 & 94.05 \\

& GeoBridge
& {95.82} & {97.77}
& {97.14} &{95.05} \\
\hline

Continual
& GeoMFD ($\mathcal{O}_1$)
& 94.71 & 95.69
& {97.15} &{94.50} \\
\hline
\end{tabular}%
\caption{Comparison with state-of-the-art methods on U-1652.
Individual methods are independently trained on U-1652. The GeoMFD
results are evaluated after completing the continual sequence
$\mathcal{O}_1=[\mathcal{D}_5,\mathcal{D}_1,\mathcal{D}_2,
\mathcal{D}_4,\mathcal{D}_3]$, where U-1652
($\mathcal{D}_5$) is introduced at the first stage.}
\label{tab:u1652_results}
\end{table}

\subsection{Performance Retention on Initial Environments}
\label{sec:supp_initial_environment}

The main comparison includes methods with publicly available implementations,
allowing all approaches to be evaluated consistently under the C-DVGL
setting. Several recent DVGL methods, including
APA-BI~\cite{zhang2025apa}, SHAA~\cite{chen2025shaa},
CDM-Net~\cite{zhou2025cdm}, and GeoBridge~\cite{song2026geobridge},
have reported results on U-1652 or SUES-200, but their public
implementations are unavailable. We therefore include their
reported results in the following dataset-specific comparisons to provide a
more comprehensive evaluation against recent DVGL methods. These experiments
also examine whether GeoMFD can retain knowledge learned from an environment
introduced at the beginning of a continual sequence. Specifically, SUES-200
and U-1652 are placed at the first stage under $\mathcal{O}_3$ and
$\mathcal{O}_1$, respectively, and are evaluated after the model has
subsequently adapted to the other four environments.

\noindent\textbf{SUES-200 as the Initial Environment.}
Table~\ref{tab:sues_all_heights} complements the main comparison by reporting
results at all four flight altitudes. Under
$\mathcal{O}_3=[\mathcal{D}_3,\mathcal{D}_4,\mathcal{D}_2,
\mathcal{D}_1,\mathcal{D}_5]$, SUES-200 is introduced at the first stage.
After adaptation to the remaining four environments, GeoMFD achieves
Drone$\rightarrow$Satellite R@1/AP scores of 99.33\%/99.46\%,
99.80\%/99.85\%, 99.82\%/99.87\%, and 99.95\%/99.96\% at
150\,m, 200\,m, 250\,m, and 300\,m, respectively, outperforming the
individually trained methods at all four altitudes. For
Satellite$\rightarrow$Drone retrieval, its R@1 remains between 98.75\% and
100.00\%, while AP ranges from 98.27\% to 98.68\%, remaining competitive
with the strongest individually trained methods. These results indicate that
GeoMFD effectively retains the knowledge learned from SUES-200 across
different flight altitudes after adapting to four subsequent environments.

\noindent\textbf{U-1652 as the Initial Environment.}
Under
$\mathcal{O}_1=[\mathcal{D}_5,\mathcal{D}_1,\mathcal{D}_2,
\mathcal{D}_4,\mathcal{D}_3]$, U-1652 is introduced at the first stage.
As shown in Table~\ref{tab:u1652_results}, GeoMFD achieves
94.71\%/95.69\% R@1/AP for Drone$\rightarrow$Satellite retrieval and
97.15\%/94.50\% for Satellite$\rightarrow$Drone retrieval after completing
the entire sequence. These results remain competitive with methods trained
exclusively on U-1652. In particular, GeoMFD achieves the highest
Satellite$\rightarrow$Drone R@1 and the second-highest AP, despite
subsequently adapting to four additional environments. This demonstrates that
the cross-view matching capability learned from U-1652 is effectively
retained during later adaptation.

The results on SUES-200 and U-1652 demonstrate that GeoMFD
consistently retains knowledge acquired from the initial environment across
different datasets, continual orders, flight altitudes, and retrieval
directions.

\begin{figure}[ht]
  \centering
  \includegraphics[width=3.2in]{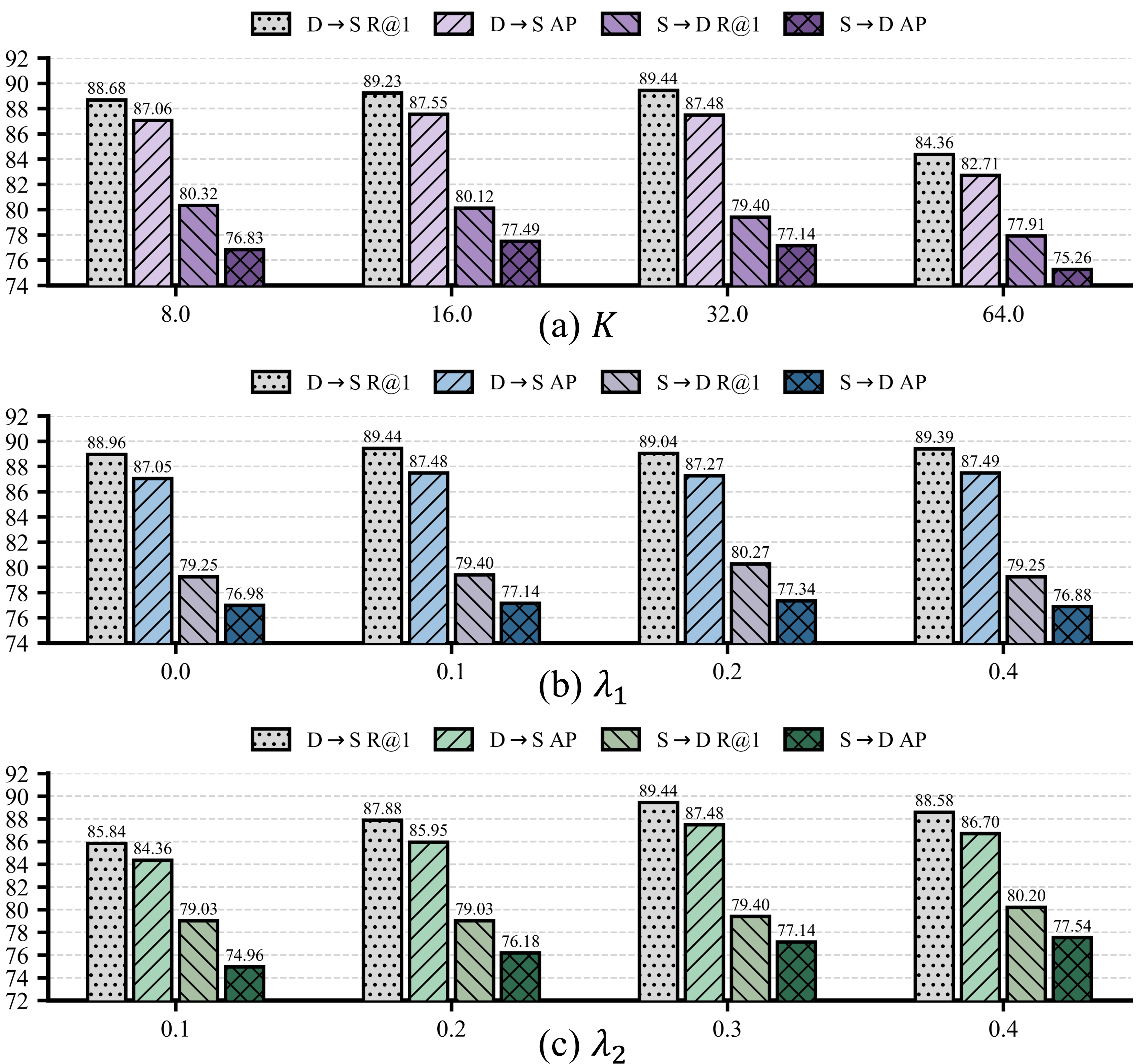}
\caption{Hyperparameter sensitivity of GeoMFD under the default continual
order $\mathcal{O}_0$. We report the average R@1 and AP over the five datasets
for both retrieval directions. (a) Number of top-$K$ hard negatives
$K$. (b) Weight $\lambda_{\mathrm{1}}$ of the view-gap closure loss.
(c) Weight $\lambda_{\mathrm{2}}$ of margin-field distillation.}
\label{fig:hyperparameter_analysis}
\end{figure}

\subsection{Hyperparameter Analysis}
\label{sec:hyperparameter_analysis}

We analyze the sensitivity of GeoMFD to three key hyperparameters, including
the number of top-$K$ hard negatives, the weight
$\lambda_{\mathrm{1}}$ of the view-gap closure loss, and the weight
$\lambda_{\mathrm{2}}$ of margin-field distillation. All results are obtained
under the default continual order $\mathcal{O}_0$ and are averaged over the
five datasets.

\noindent\textbf{Effect of the number of hard negatives $K$.}
Figure~\ref{fig:hyperparameter_analysis}(a) studies the effect of the number of
top-$K$ hard negatives. GeoMFD remains relatively stable when $K$
varies from $8$ to $32$. Increasing $K$ from $8$ to $32$ improves
Drone$\rightarrow$Satellite R@1 from $88.68\%$ to $89.44\%$, indicating that a
moderately larger hard-negative set provides more informative constraints
around the retrieval boundary. The best AP values for the two retrieval
directions are obtained at $K=16$, reaching $87.55\%$ and $77.49\%$,
respectively, while $K=32$ achieves the highest
Drone$\rightarrow$Satellite R@1. In contrast, increasing $K$ to $64$ causes a
clear performance degradation. In particular, Drone$\rightarrow$Satellite
R@1/AP decreases to $84.36\%/82.71\%$. This suggests that including too many
negatives introduces less informative or potentially unreliable teacher
relations, thereby weakening the boundary-focused property of MFD. We use
$K=32$ because it provides strong performance in both directions while
covering a sufficiently broad set of hard negative relations.

\noindent\textbf{Weight of the view-gap closure loss.}
Figure~\ref{fig:hyperparameter_analysis}(b) evaluates
$\lambda_{\mathrm{1}}$. Without view-gap closure
($\lambda_{\mathrm{1}}=0$), GeoMFD obtains R@1/AP values of
$88.96\%/87.05\%$ and $79.25\%/76.98\%$ in the two retrieval directions.
Setting $\lambda_{\mathrm{1}}=0.1$ improves all four metrics to
$89.44\%/87.48\%$ and $79.40\%/77.14\%$, confirming that explicitly
encouraging the adapter to compensate for the cross-view discrepancy is
beneficial. A larger weight of $0.2$ further improves
Satellite$\rightarrow$Drone retrieval but slightly reduces the Drone→Satellite retrieval performance. When the weight is increased to $0.4$, the
Satellite$\rightarrow$Drone metrics decrease to $79.25\%/76.88\%$. These
results show that view-gap closure acts as a useful auxiliary constraint but
should not dominate the retrieval objective. We therefore set
$\lambda_{\mathrm{1}}=0.1$ to obtain a balanced improvement across both
directions.

\noindent\textbf{Weight of margin-field distillation.}
Figure~\ref{fig:hyperparameter_analysis}(c) examines
$\lambda_{\mathrm{2}}$. Increasing the weight from $0.1$ to $0.3$
consistently improves Drone$\rightarrow$Satellite R@1/AP from
$85.84\%/84.36\%$ to $89.44\%/87.48\%$. Over the same range,
Satellite$\rightarrow$Drone AP increases from $74.96\%$ to $77.14\%$.
These improvements demonstrate that an adequate distillation strength is
necessary to preserve historical positive-versus-hard-negative relations.
Further increasing the weight to $0.4$ improves
Satellite$\rightarrow$Drone R@1/AP to $80.20\%/77.54\%$, but decreases
Drone$\rightarrow$Satellite R@1/AP to $88.58\%/86.70\%$. This trade-off
indicates that overly strong distillation restricts the model's plasticity
during adaptation to the current environment. We consequently choose
$\lambda_{\mathrm{2}}=0.3$, which provides the best
Drone$\rightarrow$Satellite performance while maintaining competitive
Satellite$\rightarrow$Drone accuracy.


\end{document}